\documentclass{article} 
\usepackage{iclr2026_conference,times}

\usepackage{hyperref}
\usepackage{url}
\usepackage{graphicx}

\usepackage{url}
\usepackage{caption}
\usepackage{subcaption}
\usepackage{mathtools}

\usepackage{multicol}
\usepackage{multirow}
\usepackage{booktabs}       
\usepackage{makecell}
\definecolor{lightgrey}{gray}{0.9}
\usepackage{xcolor}
\usepackage{pifont}
\usepackage{amssymb}
\usepackage{bbding}
\usepackage{xcolor,colortbl}
\definecolor{minetable1colorx}{rgb}{0.75, 0.75, 0.75}
\newcommand{\mineyes}{{\scriptsize \CheckmarkBold}}
\newcommand{\mineno}{{\scriptsize \textcolor{minetable1colorx}{\XSolidBrush}}}
\usepackage{algorithm}
\usepackage{algorithmicx}
\usepackage{algpseudocode}
\usepackage{algorithm}
\usepackage{enumitem}
\definecolor{BestBG}   {HTML}{FFCDCB}
\definecolor{SecondBG} {HTML}{FFE3CF}
\definecolor{ThirdBG}  {HTML}{CAE5FF}

\newcommand{\best}[1]{\cellcolor{BestBG}\textbf{#1}}
\newcommand{\second}[1]{\cellcolor{SecondBG}{#1}}
\newcommand{\third}[1]{\cellcolor{ThirdBG}{#1}}

\newcommand{\redregion}[1]{%
  \begingroup\setlength{\fboxsep}{0.2ex}\colorbox{BestBG}{#1}\endgroup}
\newcommand{\orangeregion}[1]{%
  \begingroup\setlength{\fboxsep}{0.2ex}\colorbox{SecondBG}{#1}\endgroup}
\newcommand{\blueregion}[1]{%
  \begingroup\setlength{\fboxsep}{0.2ex}\colorbox{ThirdBG}{#1}\endgroup}

\newcommand\blfootnote[1]{%
  \begingroup
  \renewcommand\thefootnote{*}\footnote{#1}%
  \addtocounter{footnote}{-1}%
  \endgroup
}

\newcommand{\shortname}{DrivingGen }

\title{DrivingGen: A Comprehensive Benchmark for Generative Video World Models in Autonomous Driving}

\author{
\centerline{\large
    Yang Zhou\textsuperscript{1}\textsuperscript{\thanks{Equal contribution.}}
\quad Hao Shao\textsuperscript{2}\textsuperscript{\blfootnote{Equal contribution.}}
\quad Letian Wang\textsuperscript{1}
} 
\\ \\ 
\centerline{\large
\textbf{Zhuofan Zong\textsuperscript{2}}
\quad \textbf{Hongsheng Li\textsuperscript{2}}
\quad \textbf{Steven L. Waslander\textsuperscript{1}}} 
\\ \\
\centerline{\large 
\textsuperscript{1}{\ University of Toronto} \quad \textsuperscript{2}{\ CUHK MMLab}}
\\ \\
\centerline{\large
    Project Website: \href{https://drivinggen-bench.github.io/}{\textcolor{magenta}{https://drivinggen-bench.github.io/}}
}
}

\iclrfinalcopy 
\begin{document}

\maketitle

\begin{abstract}

Video generation models, as one form of world models, have emerged as one of the most exciting frontiers in AI, promising agents the ability to imagine the future by modeling the temporal evolution of complex scenes. 
In autonomous driving, this vision gives rise to driving world models: generative simulators that imagine ego and agent futures, enabling scalable simulation, safe testing of corner cases, and rich synthetic data generation.
Yet, despite fast-growing research activity, the field lacks a rigorous benchmark to measure progress and guide priorities. Existing evaluations remain limited: generic video metrics overlook safety-critical imaging factors; trajectory plausibility is rarely quantified; temporal and agent-level consistency is neglected; and controllability with respect to ego conditioning is ignored. Moreover, current datasets fail to cover the diversity of conditions required for real-world deployment.
To address these gaps, we present DrivingGen, the first comprehensive benchmark for generative driving world models. DrivingGen combines a diverse evaluation dataset curated from both driving datasets and internet-scale video sources, spanning varied weather, time of day, geographic regions, and complex maneuvers, with a suite of new metrics that jointly assess visual realism, trajectory plausibility, temporal coherence, and controllability. Benchmarking 14 state-of-the-art models reveals clear trade-offs: general models look better but break physics, while driving-specific ones capture motion realistically but lag in visual quality.
DrivingGen offers a unified evaluation framework to foster reliable, controllable, and deployable driving world models, enabling scalable simulation, planning, and data-driven decision-making.

\end{abstract}

\section{Introduction}

Driven by scalable learning techniques, generative video models have made remarkable progress in recent years, enabling the synthesis of high-fidelity videos across diverse scenes and motions. These models suggest a promising path toward “world models” – predictive simulators capable of imagining the future, which can support planning, simulation, and decision-making in complex, dynamic environments.
Inspired by this vision, there has been an accelerating surge in developing driving world models: generative models specialized for predicting future driving scenarios. Given an initial scene and optional conditions (\textit{e.g.}, text prompts, driving actions), a driving world model predicts both the ego-vehicle’s future movements and the evolution of surrounding agents' trajectories. Such models enable closed-loop simulation and synthetic data generation, reducing reliance on real-world data and offering a promising means to explore out-of-distribution scenarios safely~\citep{gao2024vista,hassan2024gemgeneralizableegovisionmultimodal,mousakhan2025orbisovercomingchallengeslonghorizon,li2025driversenavigationworldmodel,wang2025deployable,zhou2025smartpretrain}. 
Driving world models are also tightly coupled with end-to-end autonomous driving systems, where errors in predicted future scenes and trajectories can directly lead to unsafe decisions~\citep{shao2023safety, shao2024lmdrive, shao2023reasonnet, wang2023efficient}.

\begin{table*}[!h]
\centering
\resizebox{0.95\linewidth}{!}{
\begin{tabular}{l|ccccc}
\toprule
\multirow{3}{*}{\textbf{Method / Benchmark}} & \multicolumn{5}{c}{\textbf{Evaluation Metrics}} \\
\cmidrule{2-6}
 & \multirow{2}{*}{\textbf{Distribution}} 
 & \multirow{2}{*}{\textbf{Quality}}
 & \multirow{2}{*}{\textbf{\makecell[c]{Temporal\\Consistency}}}
 & \multirow{2}{*}{\textbf{Alignment}}
 & \multirow{2}{*}{\textbf{\makecell[c]{Downstream\\Task}}} 
 \\
 \\
 \cmidrule{1-6}
 VBench~\citep{huang2023vbenchcomprehensivebenchmarksuite}  & \mineno & \mineyes & \mineyes & \mineno & \mineno \\
 WorldModelBench~\citep{li2025worldmodelbenchjudgingvideogeneration}  & \mineno & \mineyes & \mineyes & Instruction & \mineno \\
 WorldScore~\citep{duan2025worldscoreunifiedevaluationbenchmark} & \mineno & \mineyes & \mineyes & Traj. & \mineno \\
  \midrule
Vista~\citep{gao2024vista} & Visual & Human eval & \mineno & Traj. & \mineno \\
 GEM~\citep{hassan2024gemgeneralizableegovisionmultimodal} & Visual & Human eval & Human eval, Agent & Traj. & \mineno \\
 Doe-1~\citep{zheng2024doe1closedloopautonomousdriving} & Visual & \mineno & \mineno & \mineno & VQA, Planning \\
Drivingdojo~\citep{wang2024drivingdojodatasetadvancinginteractive} & Visual & \mineno & \mineno & Traj. & \mineno \\
 Driverse~\citep{li2025driversenavigationworldmodel} & Visual & \mineno & \mineno & Traj. & \mineno \\
UniFuture~\citep{liang2025seeingfutureperceivingfuture} & Visual & \mineno & \mineno & \mineno & Perception \\
 VaViM~\citep{bartoccioni2025vavimvavamautonomousdriving} & Visual & \mineno & \mineno & \mineno & Segmentation \\ 
GAIA-2~\citep{russell2025gaia2controllablemultiviewgenerative} & Visual & \mineno & Visual, Agent & \mineno & \mineno \\
 ReSim~\citep{yang2025resimreliableworldsimulation} & Visual & \mineno & \mineno & Traj. & Planning \\
 \midrule
 ACT-Bench~\citep{arai2024actbenchactioncontrollableworld} & \mineno & \mineno & \mineno & Instruction, Traj. & \mineno \\
 \cellcolor{lightgrey}\shortname (Ours) & \cellcolor{lightgrey}Visual, Traj. & \cellcolor{lightgrey}Visual, Traj.& \cellcolor{lightgrey}Visual, Agent, Traj. & \cellcolor{lightgrey}Traj. & \cellcolor{lightgrey}\mineno \\
\bottomrule

\end{tabular}
}
\caption{Comparison of existing video benchmarks, driving world models, and driving video benchmarks. ``\mineno" indicates the missing metrics, and ``\mineyes" signifies that the evaluation is comprehensive. ``Visual'', ``Agent'' and ``Traj.'' represent evaluation of images or videos, surrounding agents and vehicles' trajectories, respectively.}
\label{tab:world_model_evaluation}
\end{table*}
 
While a vibrant exploration of a wide range of approaches for driving world models is underway, a well-designed benchmark – which not only measures progress but also guides research priorities and shapes the trajectory of the entire field – has not yet emerged. Current evaluations fail to fully capture the unique requirements of the driving domain, and are limited in several ways.  
1) \textit{Visual Fidelity} First, most benchmarks rely on distribution-level metrics such as Fréchet Video Distance (FVD) to assess video realism, and some adopt human-preference-aligned models (e.g., vision-language models) to score visual quality or semantic consistency. However, driving imposes unique constraints on imaging: sensor artifacts, glare, or other corruptions can have critical safety implications that general video metrics fail to capture.  
2) \textit{Trajectory Plausibility} Second, the ego-motion trajectories underlying the generated videos are crucial. High-quality video generation in driving must produce trajectories that are natural, dynamically feasible, interaction-aware, and safe—properties that go beyond mere visual realism.  
3) \textit{Temporal and Agent-Level Consistency} Third, temporal consistency is crucial for driving, where surrounding objects directly impact safety and decision-making. Prior benchmarks often focus on scene-level consistency but neglect agent-level consistency, such as abrupt appearance changes or abnormal disappearances of agents—imperfections that can severely compromise the realism and reliability of driving simulations.
4) \textit{Motion Controllability} 
Finally, for ego-conditioned video generation, it is critical to assess whether the generated motion faithfully follows the conditioning trajectory. This aspect of controllability is largely overlooked in existing benchmarks, yet it is essential for safe planning and reliable closed-loop driving, where misalignment can lead to catastrophic consequences.

Another major limitation in existing benchmarks for driving world models is the lack of diversity along crucial dimensions essential for real-world deployment. 1) First, \textit{Weather and Time of Day} coverage is heavily skewed: datasets like nuScenes~\citep{caesar2020nuscenesmultimodaldatasetautonomous} are dominated by clear-weather, daytime driving, leaving rare but safety-critical conditions (night, snow, fog) underrepresented. 2) Second, \textit{Geographic Coverage} is limited, often confined to a few cities or countries, which restricts evaluation across varied scene appearance and with local traffic rules. 3) Third, \textit{Driving Maneuvers and Interactions} rarely capture the full diversity of agent behaviors and complex multi-agent dynamics, such as pedestrians waiting at crosswalks, aggressive driver cut-ins, or dense traffic scenarios~\citep{wang2021socially}. This lack of diversity makes it difficult to assess whether generative models can handle the wide range of scenarios encountered in real-world driving, undermining their reliability and safety for large-scale deployment.

\begin{figure}
    \centering
    \includegraphics[width=0.85\linewidth]
    {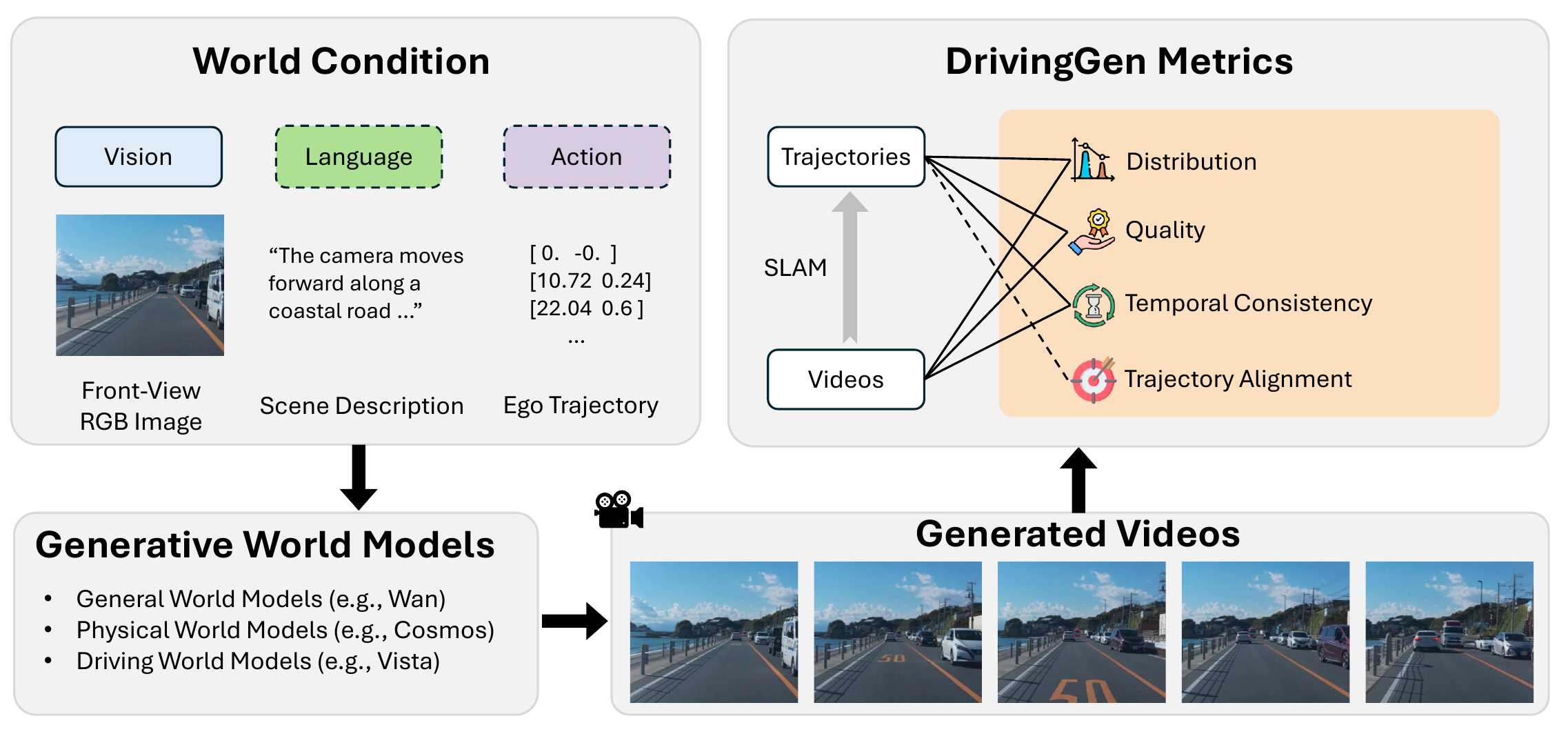}
    \caption{Overview of our \shortname benchmark. Video models take vision, and optional language/action as inputs to generate videos.  The generated videos are then passed into our evaluation suite. 
    Four comprehensive and novel sets of metrics for both videos and trajectories (distribution, quality, temporal consistency, and trajectory alignment) are introduced to evaluate world models. 
    }
    \label{fig:pipe}
\end{figure}

To address the above gaps, this work proposes DrivingGen, a comprehensive benchmark for generative world models in the driving domain with a diverse data distribution and novel evaluation metrics. DrivingGen evaluates models from both a visual perspective (the realism and overall quality of generated videos) and a robotics perspective (the physical plausibility, consistency and accuracy of generated trajectories). Our benchmark makes the following key contributions:

\textbf{Diverse Driving Dataset}. We present a new evaluation dataset that captures diverse driving conditions and behaviors. Unlike prior datasets biased toward sunny, daytime urban scenes, ours includes varied weather (rain, snow, fog, floods, sandstorms), times of day (dawn, day, night), global regions (North America, Europe, Asia, Africa, etc.), and complex scenarios (dense traffic, sudden cut-ins, pedestrian crossings). This diversity enables more robust and unbiased evaluation of generative models under realistic driving distributions. Besides, considering that inference for video generation is generally time-consuming, we carefully limit the number of samples to 400 to ensure efficient testing and iteration, achieving a balance between efficiency and meaningful evaluation.

\textbf{Driving-Specific Evaluation Metrics}. We introduce a novel suite of multifaceted metrics specifically designed for driving scenarios. These include distribution-level measures for both video and trajectory outputs, 
quality metrics that account for human perceptual quality, driving-specific imaging factors (such as illumination flicker, motion blur, etc.), temporal consistency checking at both the scene level and individual agent level (\textit{e.g.}, appearance discrepancy or unnatural disappearances in videos), and trajectory realism metrics that evaluate kinematic feasibility and alignment to intended paths (\textit{e.g.}, smoothness, physical plausibility, and accuracy in following a given route). Together, these metrics provide a comprehensive 4-dimensional evaluation along distribution realism, visual quality, temporal coherence, and control/trajectory fidelity – covering aspects that generic metrics or single-number scores fail to capture.

\textbf{Extensive Benchmarking and Insights}. We benchmark 14 generative world models on DrivingGen spanning three categories – general video world models, physics-based world models, and driving-specialized world models. This evaluation, the first of its kind in the driving domain, reveals important insights and open challenges. For example, we find that certain general world models produce visually appealing traffic scenes yet break physical consistency in vehicle motion, and some driving-specific models excel in trajectory accuracy but lag in image fidelity. By analyzing performance across our metrics, we reveal the strengths and failure modes of each approach, offering insights for future research. All components of DrivingGen—dataset and evaluation code—are publicly released to support reproducible research and advance realistic driving simulation.

\begin{figure}
  \centering

  \begin{subfigure}{\linewidth}
    \centering
    \includegraphics[width=0.99\linewidth]{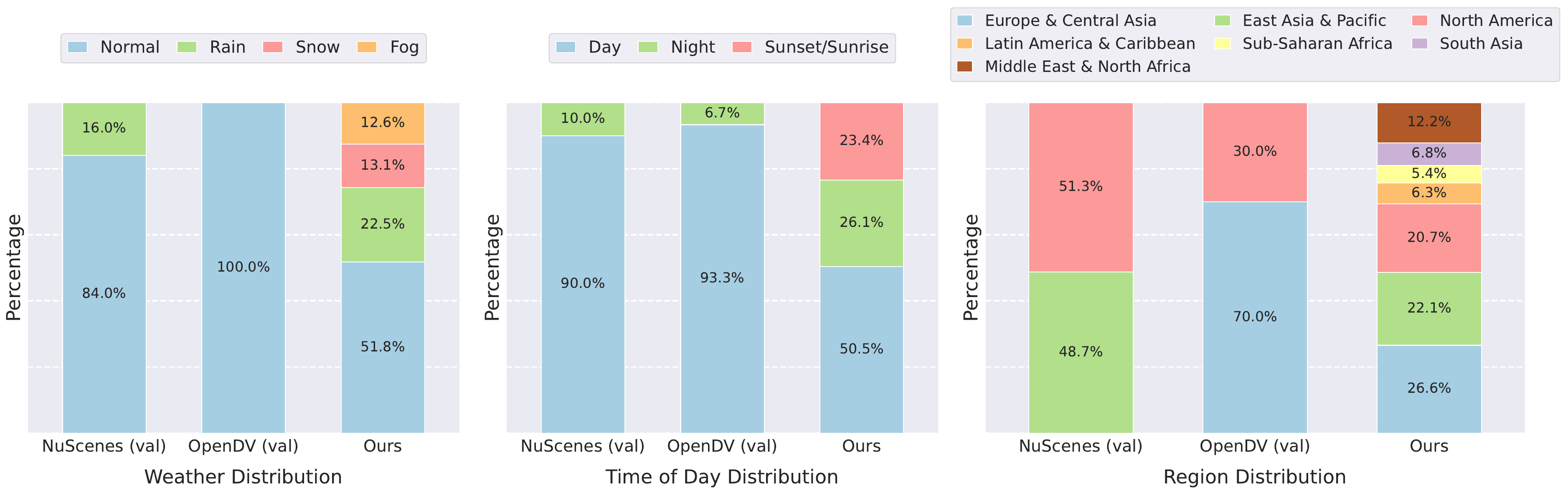}
    \caption{Weather, time of day, and region distribution between existing datasets and ours.}
    \label{fig:distribution}
  \end{subfigure}


  \begin{subfigure}{\linewidth}
    \centering
    \includegraphics[width=0.95\linewidth]{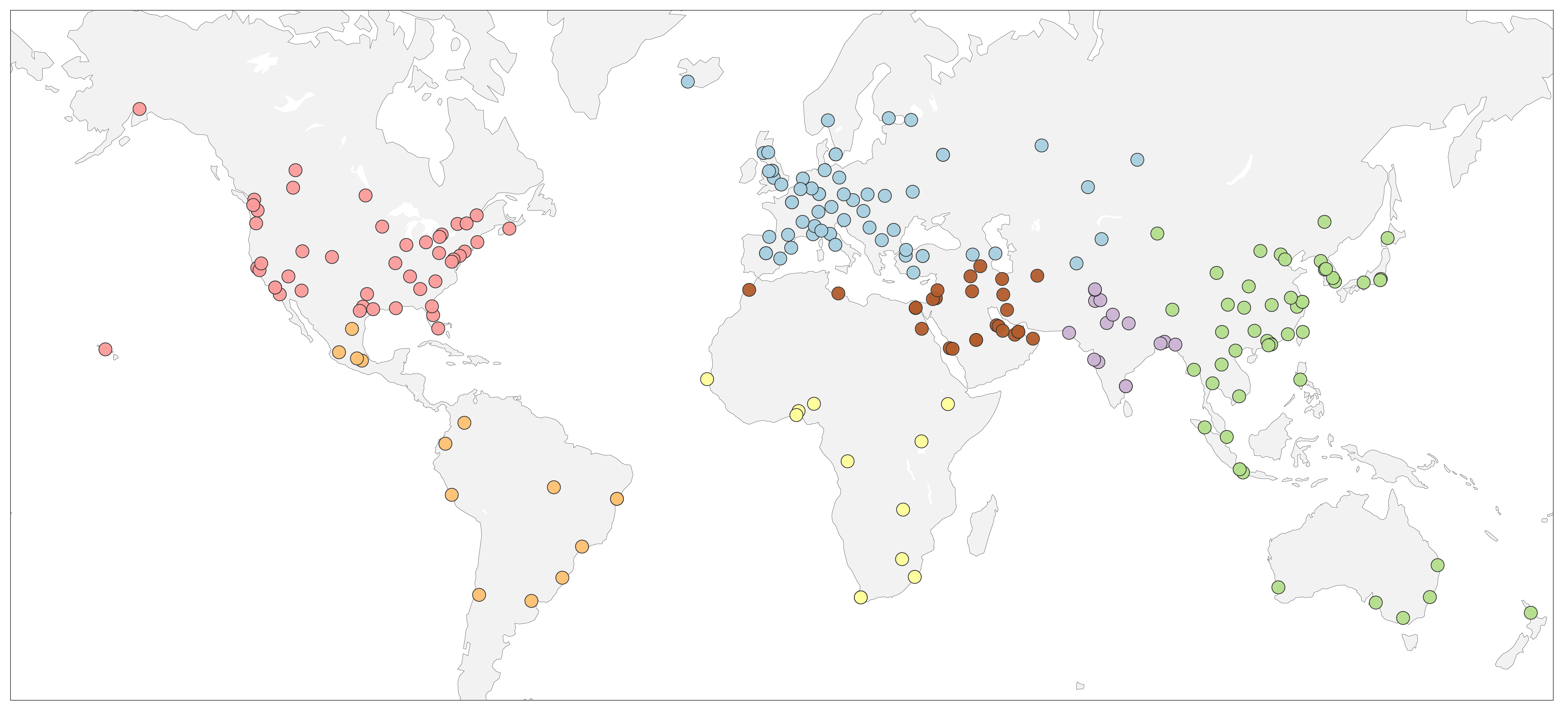}
    \caption{Specific driving locations inside each region of our dataset.}
    \label{fig:distribution_2}
  \end{subfigure}


  \begin{subfigure}{\linewidth}
    \centering
    \includegraphics[width=0.95\linewidth]{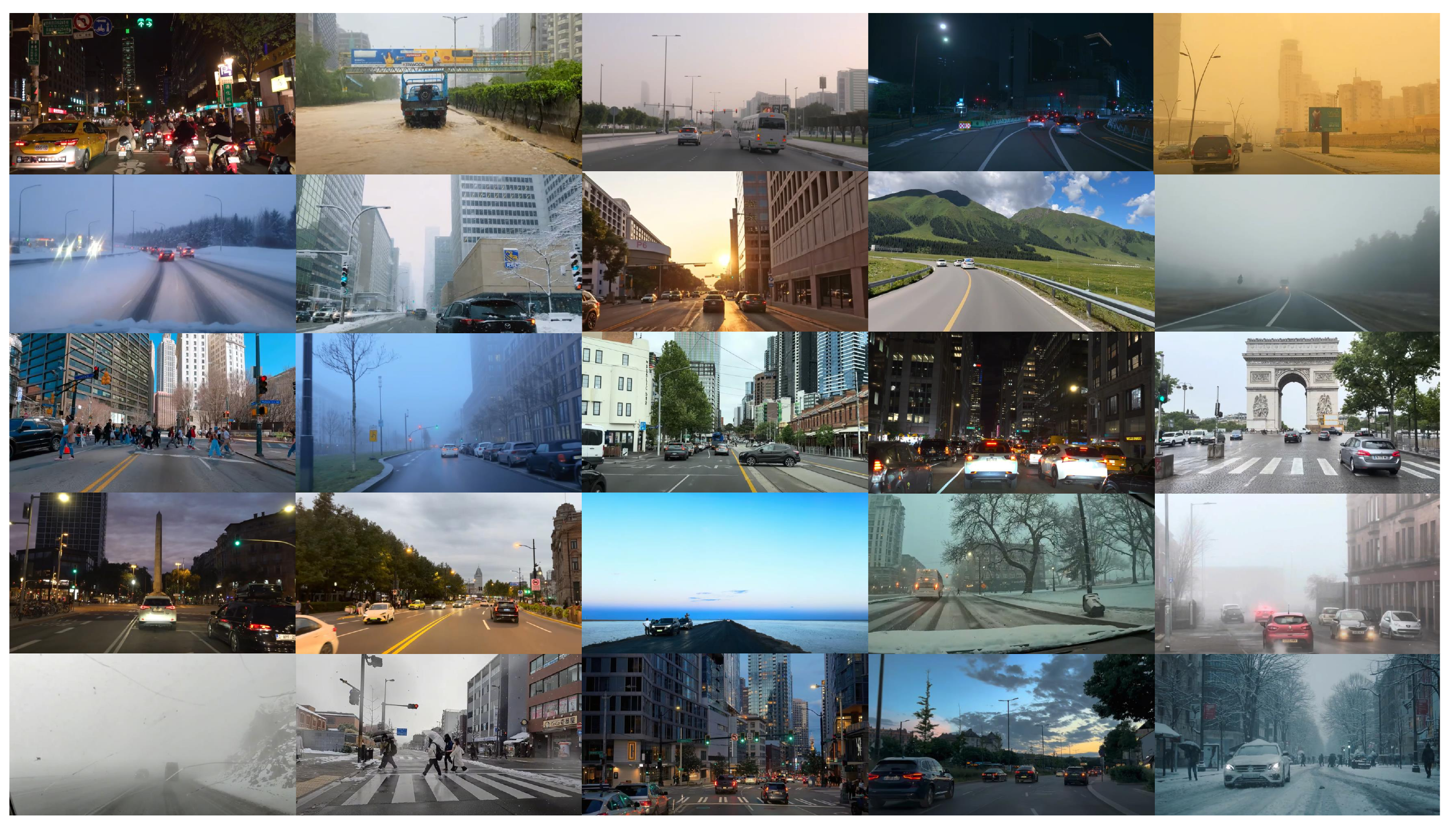}
    \caption{
    Representative examples in our benchmark, which covers diverse scenarios such as dense city traffic at night, unusual weather (\textit{e.g.}, fog, flood, sandstorm), and complex interactions (\textit{e.g.}, waiting for pedestrians, agents cutting in).
    }
  \end{subfigure}

  \caption{Dataset distribution and gallery in our benchmark (top to bottom).}
  \label{fig:gallery}
\end{figure}

\section{Related Works}

In this work, we focus on two primary research areas: generative world models applied to autonomous driving and benchmarks for evaluating these models. Due to space constraints, we provide a comprehensive review of the relevant literature, including recent advancements in general video generation and specific driving-world evaluations, in Appendix~\ref{app:related_works}.

\section{\shortname Benchmark}

The goal of \shortname is to establish a comprehensive benchmark to evaluate generative world models under driving-specific constraints and criteria. To achieve this, the proposed benchmark includes several key components: 1) a carefully collected dataset that is diverse in weather, time of day, regions (and their driving styles), and driving maneuvers to support reasonable evaluation; 2) multifaceted metrics that not only evaluate the video quality from a general visual perspective (\textit{e.g.}, appearance), but also from a driving and robotics perspective (\textit{e.g.}, the physical feasibility of trajectories). 
To showcase the distinguishing capability of DrivingGen, we evaluate general world models, physics-based models, and driving-specific models.  
An overview is given in Fig.~\ref{fig:pipe}, with dataset details in Sec.~\ref{sec:dataset} and metrics in Sec.~\ref{sec:metrics}.

\subsection{Benchmark Dataset}
\label{sec:dataset}

Generative video models, as a form of world models, offer a promising way to anticipate future driving scenarios, simulate rare or safety-critical events, and ultimately support planning and decision-making.
However, real-world driving unfolds under highly variable conditions, encompassing different weather, lighting, regions, and complex maneuvers.
Therefore, evaluating generative models across diverse scenarios is crucial to ensure their robustness and reliability.
To this end, the majority of existing works~\citep{gao2024vista,hassan2024gemgeneralizableegovisionmultimodal,liang2025seeingfutureperceivingfuture,wang2024drivingdojodatasetadvancinginteractive,bartoccioni2025vavimvavamautonomousdriving,wang2024distillnerf} in driving world models mainly utilize nuScenes~\citep{caesar2020nuscenesmultimodaldatasetautonomous} and OpenDV~\citep{zheng2024genadgenerativeendtoendautonomous} datasets for evaluation. However, the diversity of weather, region, time of day, and driving maneuvers in these datasets is limited and highly biases the data distribution.
For example, as shown in Fig.~\ref{fig:distribution}, over 80\% of the nuScenes validation data and 90\% of the OpenDV validation data are collected during normal sunny daytime conditions.
Additionally, the data are collected from a limited number of vehicles and locations, which further limits the comprehensiveness. Based on this observation, we curated a significantly more diverse dataset. An overview of our dataset is presented in Fig.~\ref{fig:distribution} and Fig.~\ref{fig:distribution_2}.

\textbf{Dataset Construction.} We organize our dataset into two complementary tracks, offering distinct perspectives for evaluating driving videos.

\begin{itemize}[leftmargin=1.5em,labelsep=0.5em]
    \item \textit{Open-Domain Track} is designed to evaluate models’ generalization to open-domain, diverse, unseen driving scenarios. We construct this track using Internet-sourced data spanning multiple cities and regions worldwide, ensuring broad coverage beyond the training distribution.
    \item  \textit{Ego-Conditioned Track} complements the open-domain track. While the open-domain setting evaluates generalization to diverse unseen scenarios, it does not verify whether the generated trajectories follow a specified conditioning trajectory—a property that is critical for robotics and self-driving applications. The ego-conditioned track therefore focuses on trajectory controllability, measuring how well the trajectories derived from generated videos align with the given ego-trajectory instructions. The ego trajectory is optional for model input and only provided in this track. To construct it, we aggregate data from five open-source driving datasets: Zod~\citep{zod2023} (Europe), DrivingDojo~\citep{wang2024drivingdojodatasetadvancinginteractive} (China), COVLA~\citep{covla_wacv2025} (Japan), nuPlan~\citep{karnchanachari2024learningbasedplanningthenuplanbenchmark} (US), and WOMD~\citep{sun2020scalabilityperceptionautonomousdriving} (US).

    
\end{itemize}

Each data sample in the dataset consists of three components: a front-view RGB image (vision), a scene description (language), and an optional ego trajectory (action). For each scene, we employ Qwen~\citep{qwen25vl2025} to capture descriptions of the future dynamics and camera movements within the scene. 
Given the time-consuming nature of video generation, we limit the number of samples for efficient testing and iteration, while ensuring quality and diversity. The dataset includes 400 samples—200 per track—striking a balance between efficiency and meaningful evaluation.

\textbf{Balanced Data Dsitribution}
The overall distribution of our dataset, along with a gallery of representative video examples, is shown in Fig.~\ref{fig:gallery}. To ensure meaningful evaluation, we explicitly control diversity across several dimensions:
\begin{itemize}[leftmargin=1.5em,labelsep=0.5em]

\item \textit{Weather and Time of Day.} Existing benchmarks are often dominated by, if not fully composed of, normal weather and daytime conditions. In contrast, our benchmark aims for a more balanced distribution. For the open-domain track, we limit normal weather and daytime clips to below 60\% and increase the proportion of other conditions, such as snow (13.1\%), fog (12.6\%), and night/sunset/sunrise driving (50\%), to ensure a more comprehensive evaluation. Extreme events, including sandstorms, floods, and heavy snowfall at night, are also included. A similar strategy is applied to the ego-conditioned track, where normal weather/daytime clips make up 60\% of the data, while the remainder covers diverse conditions to support trajectory controllability evaluation across different scenarios.

\item \textit{Geographic Coverage.} Prior benchmarks are often limited to a small number of cities or countries, restricting the diversity of driving scenarios. For the open-domain track, we collect data from a wide range of regions worldwide, including North America (20.7\%), East Asia \& Pacific (22.1\%), Europe \& Central Asia (26.6\%), the Middle East \& North Africa (12.1\%), Latin America \& Caribbean (6.3\%), South Asia (6.8\%) and South-Saharan Africa (5.4\%),
to ensure broad geographic coverage. For the ego-conditioned track, data are drawn from existing datasets covering North America, Asia and Europe, providing diverse driving scenarios to evaluate ego-trajectory alignment and controllability.

\item \textit{Driving Maneuvers and Interactions.} Capturing diverse driving behaviors and multi-agent interactions is critical for evaluating generative world models. For the open-domain track, scenarios include complex interactions such as waiting pedestrians at crosswalks, other agents cutting in, and dense traffic, testing the model's understanding of the driving world. For the ego-conditioned track, scenarios are similarly diverse, emphasizing multi-agent interactions and challenging conditions to evaluate controllability and alignment with ego-trajectory instructions.

\end{itemize}

\subsection{Benchmark Metrics}
\label{sec:metrics}

\begin{table*}[!t]

\def\arraystretch{1.05} 
    \centering
    \resizebox{0.95\linewidth}{!}{
    \begin{tabular}{cccc}
        \toprule
          Distribution & Quality & Temporal Consistency & Trajectory Alignment \\ 
          \midrule
          \multirow{4}{*}{\makecell[l]{1. Fréchet Video Distance (FVD)\\2. Fréchet Trajectory Distance (FTD)}} &
          \multirow{4}{*}{\makecell[l]{1. Subjective Image Quality\\2. Objective Image Quality\\3. Trajectory 
          Quality
          }} &
          \multirow{4}{*}{\makecell[l]{1. Video Consistency\\2. Agent Consistency\\3. Agent Disappearance Consistency\\4. Trajectory Consistency}} &
          \multirow{4}{*}{\makecell[l]{1. Average Displacement Error (ADE)\\2. Dynamic Time Warping (DTW)}}
           \\
        &&&\\
        &&&\\
        &&&\\
         \bottomrule
    \end{tabular}
    }
    \caption{Overview of metrics utilized in DrivingGen. 
    Definition and details are in Sec.~\ref{sec:metrics}.
    }
    \label{tab:metric}
\end{table*}

For all video models, our \shortname metrics cover three key dimensions: distribution, quality, and temporal consistency, evaluated for both videos and trajectories. We extract trajectories using 
standard PnP method within a SIFT and RANSAC scheme~\citep{Lowe2004-pe, Fischler1981-cb, 5995464}
and UniDepthV2~\citep{piccinelli2025unidepthv2}.
\textcolor{black}{We provide the details of our SLAM pipeline~\citep{7219438,schoenberger2016sfm,teed2022droidslamdeepvisualslam,qu2024implicit}, including guaranteeing that all videos reconstruct trajectories and a discussion to compare other benchmarks' trajectory reconstruction methods in Appendix~\ref{app:slam_detail}.}
For models conditioned on ego trajectories, we include a fourth dimension: trajectory alignment, measuring adherence to the input. Table~\ref{tab:metric} lists the metrics, grouped into four categories detailed below, each targeting a different aspect of video fidelity.

\subsubsection{Distribution} 

\textit{How far is the generative distribution from the data distribution?}
A common practice is to measure Fréchet Video Distance (FVD)~\citep{unterthiner2019accurategenerativemodelsvideo} on generated videos. However, our key insight is that video quality is not solely determined by visual realism—equally important, especially for self-driving and embodied agents, is the realism of the induced ego-motion. Focusing only on visual fidelity gives an incomplete picture. Therefore, we evaluate distributional closeness across both videos and trajectories, capturing complementary perspectives from visual perception and robotics.

For the video distribution, we utilize FVD to quantify the similarity between generated videos and real videos. Specifically, we follow the standardized computation protocol from the original StyleGAN-V~\citep{skorokhodov2022styleganvcontinuousvideogenerator}.
For the trajectory distribution, we introduce a novel metric, Fréchet Trajectory Distance (FTD), a distributional metric tailored for evaluating driving trajectories. 
The key requirement is a trajectory encoder that maps trajectories into a latent space suitable for measuring distributional distance. To this end, we draw from the motion prediction domain—where models themselves are generative of future trajectories, and adopt the encoder of Motion Transformer (MTR)~\citep{shi2023motiontransformerglobalintention} as our encoding model. Details of FTD computation are provided in Appendix~\ref{app:ftd}.

\subsubsection{Quality}

\textit{How good are the generated videos and trajectories?} To evaluate the fidelity of generated videos and trajectories in driving scenarios, we propose a comprehensive quality suite covering three aspects: perceptual video quality, domain-specific video quality, and trajectory quality.

\textbf{Visual Quality}. A common practice in generative video evaluation is to assess general perceptual quality with automatic, reference-free estimators aligned with human judgments. Specifically, we adopt CLIP-IQA+~\citep{wang2022exploringclipassessinglook}, which leverages CLIP’s vision-language representations to predict perceptual quality scores consistent with human subjective assessments. While effective, such subjective perceptual quality does not always align with what matters for driving, which unfolds outdoors, involves multiple agents, and occurs under real-world constraints. To additionally consider driving-specific imaging quality, we further adopt the Modulation Mitigation Probability (MMP) metric from the IEEE Automotive P2020 standard~\citep{ieee2018ieee,ieee_p2020_draft_2022}. MMP targets Pulse-Width Modulation (PWM)-induced flicker that can disrupt perception and tracking, and reports the fraction of time windows where residual temporal luminance modulation falls below a small threshold. Implementation details are in Appendix~\ref{app:obj_q}.
 
\textbf{Trajectory Quality}. 
While prior evaluations often rely on video-based scores, they typically neglect whether the underlying motions are physically and kinematically plausible.
To reduce the gap, \shortname introduces a composite, reference-free metric to assess the kinematic plausibility and ride comfort. Three individual submetrics are proposed and aggregated into a single score: 1) a comfort score penalizes extremes of longitudinal jerk, lateral acceleration, and yaw-rate, yielding a score to reward smoother, more comfortable motion; 2) a motion score that discourages under-mobility, as some trajectories barely move and stay static due to the model's weak ability; 3) a curvature score summarizes how much the path turns, discouraging zig-zags and unrealistically sharp bends.
Together, these submetrics directly target properties that affect controllability, planning, and perceived comfort. Calculation details appear in Appendix~\ref{app:traj_q}.

\subsubsection{Temporal Consistency}

\textit{How temporally consistent is the generated world?} We assess the temporal consistency of both videos and trajectories. For videos, we evaluate scene-level consistency, agent-level consistency, and explicitly emphasize abnormal agent disappearance. For trajectories, we measure the consistency of speed and acceleration over time, independent of path shape and absolute mobility.

\textbf{Video Consistency}. Existing metrics directly calculate the consistency between consecutive frames (or each frame to the first) at a fixed rate. However, it is easily hackable by generating near-static videos. To measure temporal consistency while accounting for the actual motion in the scene, we first pass the generated videos through an off-the-shelf optical flow model~\citep{wang2024searaftsimpleefficientaccurate} to compute the median optical flow magnitude per frame.
We then adaptively downsample: videos with lower motion are sampled more sparsely so that the per-step displacement becomes comparable to normal/high-speed driving.
After this, the similarity of the DINOv3~\citep{simeoni2025dinov3} features between consecutive frames of the downsampled videos is reported as the video consistency score. 
Unlike fixed-stride metrics, our approach fairly measures temporal consistency across videos with varying motion speeds, preventing static or near-static videos from obtaining artificially high scores.

\textbf{Agent Appearance Consistency}. 
Measuring only scene-level features can overlook small temporal changes in individual agents, such as shifts in color, texture, or shape, while these agents are often the key focus for driving, as they would more directly impact driving behavior and safety.
To measure the agent's temporal consistency, we therefore detect agents in the first frame, track them across the video, crop their bounding boxes, and compute consistency purely at the agent level.
We use YOLOv10~\citep{wang2024yolov10} as the detector and SAM2~\citep{ravi2024sam2, yang2024samuraiadaptingsegmentmodel} for tracking.
We measure DINOv3 feature similarity across consecutive frames and to the first frame. 

\textbf{Agent Abnormal Disappearance}.
In addition to appearance stability, agents in driving scenes must persist in a physically plausible manner. Sudden, non-physical disappearances of surrounding agents are commonly observed in generated videos, which can compromise realism and safety. \shortname quantifies this by diagnosing whether an agent’s disappearance is normal (\textit{e.g.}, leaving the field of view or being occluded) or abnormal.
We consider three key frames for each disappearing agent: the first and the last frames where the agent is visible, and the first frame after it vanishes. A vision large language model (VLM)~\citep{qwen25vl2025, shao2024visual, liu2024sphinx, zong2024mova, li2024llava, qu2025spatialvla, qu2025eo}, Cosmos-Reason1~\citep{nvidia2025cosmosreason1physicalcommonsense}, is prompted to judge disappearance based on visual and motion continuity, and the agent’s local interactions with surrounding agents. We report the percentage of videos with no abnormal disappearances as the score. Implementation Details can be found in Appendix~\ref{app:agent_missing}.

\textbf{Trajectory Consistency}. Realistic driving exhibits predictable kinematics: speed varies slowly around a cruise level and acceleration does not oscillate. To reveal this property, we compute how stable a trajectory’s velocity and acceleration are over time. The average of the two scores is taken as the overall trajectory consistency score. Trajectories that jitter, stop–go, or oscillate score low, while steady cruising with gradual changes scores high. Calculation details are provided in Appendix~\ref{app:traj_c}.

\subsubsection{Trajectory Alignment}

In addition to trajectory consistency, the alignment of the trajectories underlying the generated videos with the conditioning (ego) trajectory is also critical, especially for trajectory-grounded video generation. To assess this, we propose two complementary metrics.

\textbf{Average Displacement Error} (ADE). As a common practice, ADE measures the mean pointwise distance between the generated and input trajectories across the prediction horizon. It emphasizes local, step-by-step fidelity and is standard in motion prediction and planning.

\textbf{Dynamic Time Warping} (DTW). 
In addition to ADE, which compares trajectories at each time step, we introduce a complementary metric that captures the overall contour and shape of the trajectory. Specifically, DTW~\citep{Keogh2000ScalingUD} aligns predicted and reference trajectories via non-linear time warping and measures their path-shape discrepancy using Euclidean point-wise cost.

\section{Experiments}

\begin{table}[t]


\resizebox{\textwidth}{!}{%
\centering
\small 
\setlength{\tabcolsep}{3pt} 

\begin{tabular}{@{}l*{1}{c}|*{10}{c}@{}}

\toprule
\multirow{3}{*}{\makecell[l]{Open-Domain Track\\\\Models}} & 
\multirow{3}{*}{\makecell[c]{\\\\Size}}
& \multicolumn{2}{c}{Distribution} & 
\multicolumn{3}{c}{Quality} & 
\multicolumn{4}{c}{Temporal Consistency} &
\multirow{3}{*}{\makecell[c]{Avg. Rank}}\\ \cmidrule(lr){3-4}
\cmidrule(lr){5-7} \cmidrule(lr){8-11}
& & \rotcell{FVD} & \rotcell{FTD} &
\rotcell{Subjective\\Quality} & 
\rotcell{Objective\\Quality} & 
\rotcell{Trajectory\\Quality} & 
\rotcell{Video\\Consist} & 
\rotcell{Agent\\Consist} & 
\rotcell{Agent\\Missing} &
\rotcell{Trajectory\\Consist} & \\
\midrule
Kling 2.1{*} &  -  &    {693.4}        & \best{26.73} &
\best{0.5538}  & {0.8018} & \second{0.6438} & \second{0.8945} & \third{0.7981}& \third{0.9442} & \best{0.5377}&  1\\
Gen-3 Alpha Turbo{*}  &  -      &    {801.0}     & {93.50} &
\second{0.5456}  & {0.8378} & \best{0.6535} & \third{0.8900} & \best{0.8170} & \second{0.9495} & \second{0.4788}&  2\\
\midrule
LTX-Video &13B        &  {648.2} & \third{31.29}&
{0.5215} & {0.8288}&{0.5562} & {0.8851} & {0.7449}& {0.8977} & {0.4517} & 3\\
Wan2.2-I2V & 14B  &  \second{609.0}   & {63.86}&
\third{0.5348} &{0.6396} & {0.5983}& {0.8883} & {0.7514} & {0.9128} & \third{0.4639}&  4\\
HunyuanVideo-I2V & 13B  &  {957.5}  & \second{30.95} & 
{0.4921} & {0.7207}& {0.4613}& {0.8821} & \second{0.8008} & {0.9306} & {0.4157} & 5\\
SkyReels-V2-I2V &14B       &   {876.0}     &{52.93} &
{0.5134}  & {0.7432}& {0.4799}& {0.8776} & {0.7329} & {0.9078}  & {0.4326}  & 7\\
CogVideoX & 5B       &  \third{621.2}  & {236.7} &
{0.4932} & {0.6802}& {0.3856}& {0.8211} &{0.7581} & {0.7661} & {0.2949}&  12\\
\midrule
Cosmos-Predict2 & 14B       &  \best{524.1} & {83.20}&
{0.4931} & {0.7568}&{0.5990} & {0.8597} &{0.5912} & {0.8657} &{0.3997} & 8\\
Cosmos-Predict1 & 14B         & {821.1}  & {81.22}& 
{0.5083} & {0.7207}&{0.2723} & {0.8429} & {0.6789} & {0.8796} &{0.2631}  & 13\\
\midrule
Vista & 2.5B      &  {675.7}  & 54.66 &
{0.4340} & \second{0.8468}&\third{0.6030} & {0.8565} & {0.6357} & {0.8211} & {0.4040}& 6\\
VaViM & 1.2B      &    {1446.6}   & {449.2}&
{0.4691} & \second{0.8468}&{0.3118} & \best{0.9159} & {0.7721} & \best{0.9752} & {0.0914}& 9\\
UniFuture & 3.0B       &   {774.3}   & {50.66}&
{0.4206} & \best{0.9054}&{0.4507} & {0.8799} & {0.5373} & {0.8310} &{0.3858} &10\\
GEM & 2.1B       & {770.1} & {147.1} &
{0.5168} &{0.8423} &{0.5398} & {0.8176} & {0.6099} & {0.7788} & {0.3392} & 11\\
Drivingdojo &  2.3B              & {810.4} &  {126.74} &
{0.4202} & {0.8333}&{0.4511} & {0.8480} & {0.6256} & {0.8303} & {0.2739} & 14\\
\bottomrule
\end{tabular}%
}

\resizebox{\textwidth}{!}{%
\centering
\small 
\setlength{\tabcolsep}{3pt} 

\begin{tabular}{@{}l*{1}{c}|*{12}{c}@{}}
\toprule
\multirow{3}{*}{\makecell[l]{Ego-Conditioned Track\\\\Models}} &
\multirow{3}{*}{\makecell[c]{\\\\Size}} 
& \multicolumn{2}{c}{Distribution} & 
\multicolumn{3}{c}{Quality} & 
\multicolumn{4}{c}{Temporal Consistency} &
\multicolumn{2}{c}{Trajectory Alignment} & 
\multirow{3}{*}{\makecell[c]{Avg. Rank}}\\ \cmidrule(lr){3-4}
\cmidrule(lr){5-7} \cmidrule(lr){8-11} \cmidrule(lr){12-13} 
& & \rotcell{FVD} & \rotcell{FTD} &
\rotcell{Subjective\\Quality} & 
\rotcell{Objective\\Quality} & 
\rotcell{Trajectory\\Quality} & 
\rotcell{Video\\Consist} & 
\rotcell{Agent\\Consist} & 
\rotcell{Agent\\Missing} &
\rotcell{Trajectory\\Consist} &
\rotcell{ADE} & 
\rotcell{DTW} & \\
\midrule
Kling 2.1{*} & -   &    {320.5}        & \second{23.74} &
\second{0.5468}  &  {0.7838} & \best{0.6860} & \second{0.8929} & \third{0.8186}& \second{0.9712}  &  \best{0.5430}& {29.97}&{2310} & 1 \\
Gen-3 Alpha Turbo{*}  & -        &    {555.9}     & \third{24.72} &
\best{0.5740}  & \third{0.8604} & \second{0.6770} & {0.8747} & {0.7986}& \third{0.9466}  & \third{0.4800}& {33.39}&{2749} & 3 \\
\midrule
Wan2.2-I2V & 14B  &  \best{194.4}   & {29.56} &
\third{0.5084} & {0.6982}& \third{0.6419}& \third{0.8821} & {0.7561}& {0.9034} & \second{0.4849} & {27.39}&{1901} & 2\\
LTX-Video & 13B        &  {378.1} & {61.09} &
{0.4895} & \third{0.8604}& {0.5464}& {0.8705} & {0.7708}& {0.9020}  &{0.4442} & {32.12}& {2505} & 6\\
HunyuanVideo-I2V & 13B  &  {532.9}  & \best{21.18} & 
{0.4741} & {0.6847}& {0.5542}& {0.8792} & \second{0.8240}& {0.9415}  & {0.4771} & {33.80}& {2794} & 7\\
CogVideoX & 5B       &  \third{307.1}  & {166.6} &
{0.4884} & {0.6937}& {0.4252}& {0.8167} & {0.7541} & {0.8981}  & {0.3783} &{32.67} & {2413} & 10\\
SkyReels-V2-I2V & 14B       &   {428.2}     & {57.02} &
{0.4764}  & {0.6622}& {0.5028}& {0.8661} & {0.7208}& {0.875}  &{0.4322} & {31.54}&{2594} & 11\\
\midrule
Cosmos-Predict2 & 14B       &  \second{260.5} & {56.26}&
{0.4756} & {0.8198}& {0.6424} & {0.8428} & {0.6707}& {0.8986}  & {0.4108} & \third{22.38}&\third{1490} & 4\\
Cosmos-Predict1 & 14B         & {345.2}  & {34.96} & 
{0.4783} & {0.7505}& {0.3761}& {0.8229} & {0.7423}& {0.7961}  &{0.3343} & {34.47}& {3084} & 13\\
\midrule
Vista & 2.5B      &  {392.8}  & {27.33} &
{0.4146} & {0.8198}& {0.6047}& {0.8741} & {0.6417}& {0.8676} & {0.4366} & \best{19.70}&\best{1216} & 5\\
UniFuture & 3.0B       &   {654.6}   & {37.17} &
{0.4006} & \best{0.9685}& {0.5353}& {0.8759} & {0.5525}& {0.8759} & {0.4165}& \second{20.21}&\second{1352} & 8\\
VaViM & 1.2B      &    {1222}   & {103.6} &
{0.4910} & \second{0.8694}& {0.1936}& \best{0.9428} & \best{0.8290}& \best{0.9725}  &{0.0984} & {41.92} & {3863} & 9\\
Drivingdojo & 2.3B              & {586.5} & {35.73} &
{0.4264} & {0.8198}& {0.4131}& {0.8419} & {0.6940}& {0.8439} & {0.2776} & {25.50}& {2142} & 12\\
GEM & 2.1B       & {579.9} & {97.70} &
{0.4484} & {0.8018}& {0.5085}& {0.7886} &{0.6180} & {0.7463} & {0.2983}& {25.73} & {1982} & 14\\
\bottomrule
\end{tabular}%
}
\caption{\textbf{Evaluation results of 14 generative world models on our benchmark.} Best results are in \redregion{red region}, second best are in \orangeregion{orange region}, and third best are in \blueregion{blue region}. ``*’’ indicates commercial closed-source models. Models fall into four categories: closed-source, open-source general video models, physical-world models, and driving-specific models.
}
\label{tab:evaluation_results}
\end{table}

\textbf{Evaluation Setup.} We evaluate 14 competitive generative world models on DrivingGen, spanning three categories. 1) First, we include 7 general video world models, comprising two commercial closed-source models, Gen-3~\citep{runway2024gen3} and Kling~\citep{kuaishou2024kling}, and five well-known open-source models: CogVideoX~\citep{yang2024cogvideox}, Wan~\citep{wan2025wanopenadvancedlargescale}, HunyuanVideo~\citep{kong2024hunyuanvideo}, LTX-Video~\citep{hacohen2024ltxvideo}, and SkyReels~\citep{chen2025skyreelsv2infinitelengthfilmgenerative}. 2) Second, we evaluate 2 physical world models that are developed specifically for the physical robotics domain, Cosmos-Predict1~\citep{agarwal2025cosmos} and Cosmos-Predict2~\citep{cosmospredict2_2025}. 3) Third, we assess 5 driving-specific world models: Vista~\citep{gao2024vista}, DrivingDojo~\citep{wang2024drivingdojodatasetadvancinginteractive}, GEM~\citep{hassan2024gemgeneralizableegovisionmultimodal}, VaViM~\citep{bartoccioni2025vavimvavamautonomousdriving}, and UniFuture~\citep{liang2025seeingfutureperceivingfuture}. All models are evaluated on a prediction horizon of 100 frames. 
\textcolor{black}{We report the time and resource cost for our DrivingGen benchmark in Appendix~\ref{app:time}.}

\subsection{Observations and Challenges}

Table~\ref{tab:evaluation_results} presents the results. \textcolor{black}{We provide the full table of metrics in a transparent way to evaluate the models comprehensively, and the average rank serves as a quick summary but not a definitive score.} 
We also show that our results align well with human judgement, by calculating the Spearman’s correlation coefficient (see details in Appendix~\ref{app:human_align}.) In the following, we will discuss key findings from our results.

\textbf{Closed-source models lead in visual quality and overall ranking.} Across both tracks, closed-source models consistently occupy the top positions, achieving strong perceptual scores and maintaining stable agent behavior. They rarely exhibit abnormal object disappearance and generally preserve scene coherence over time, demonstrating robust overall world generation capabilities.


\textbf{Top open-source general world models are competitive on specific metrics.} Several open-source models approach or match the closed-source leaders on individual dimensions. For example, CogVideoX and Wan achieve strong video distributional realism (low FVD) across both tracks, suggesting that open-source models can excel in targeted aspects even if they do not lead overall.


\textbf{No single model excels in both visual realism and trajectory fidelity.} We observe distinct “personas”: some models achieve high visual quality but only moderate trajectory adherence and per-agent consistency, while driving-specialized models accurately follow commanded paths with physically plausible motion (low ADE/DTW) yet underperform in visual fidelity, exhibiting noticeable artifacts. Currently, no model successfully combines strong photorealism with precise, physically consistent motion, highlighting a key frontier for driving world generation.

\textbf{Trajectory alignment remains limited, revealing substantial gaps.} Under ego-trajectory conditioning, models exhibit significant ADE/DTW errors, indicating poor adherence to commanded paths. This can stem from two main factors: 1) artifacts in the generated videos (\textit{e.g.}, texture repetition, blur, unstable geometry) that impair SLAM-based trajectory recovery, and 2) imperfect motion generation, where the model itself fails to follow the intended trajectory. These observations highlight that both video fidelity and trajectory modeling need further improvement.

\textbf{DrivingGen exposes failure modes hidden from prior single metric.}
Existing benchmarks often rely solely on distribution-level metrics such as FVD to evaluate generated driving videos. 
While useful for assessing overall distribution similarity, good FVD/FTD alone does not necessarily imply plausible driving—videos can appear distribution-close yet exhibit stop–go jitter, identity drift, or non-physical disappearances.
Similarly, high objective quality (e.g., low flicker) can coexist with poor subjective quality or unstable agent behavior. By jointly reporting distribution, perceptual quality, temporal consistency, and trajectory alignment, DrivingGen exposes these hidden failure modes and highlights precisely where each model falls short.

\section{Conclusion}
This work introduces DrivingGen, a comprehensive benchmark designed to evaluate generative world models for autonomous driving. DrivingGen integrates a diverse dataset spanning varied weather, time of day, global regions, and complex driving maneuvers with a multifaceted metric suite that jointly measures visual realism, trajectory plausibility, temporal coherence, and controllability. By benchmarking a broad spectrum of state-of-the-art models, DrivingGen reveals critical trade-offs among visual fidelity, physical consistency, and controllability, providing clear insights into the strengths and limitations of current approaches. The benchmark establishes a unified and reproducible framework that can guide the development of reliable and deployment-ready driving world models, fostering progress toward safe and scalable simulation, planning, and decision making in autonomous driving.

\textcolor{black}{
\section{Future Work and Limitations}}

\textcolor{black}{As DrivingGen is the first comprehensive benchmark for generative world models in autonomous driving, several intriguing ideas can be explored further in follow-up work.}

\textcolor{black}{\textbf{Expanding More Meaningful Data.} Currently, we collect 400 data samples (from the web and aggregated from existing driving datasets) to balance efficiency and practicality, because generating and evaluating videos is resource-intensive. With this limited number, we may not fully cover the long tail of driving scenarios. In future expansions, scaling up the dataset is an exciting future direction. As generative models become faster and datasets become more readily available, scaling up to thousands of clips is feasible and will further improve long-tail coverage.}

\textcolor{black}{\textbf{Interactive and Closed-Loop Simulation.} Ensuring reliable closed-loop performance (\textit{e.g.}, for safe planning) is crucial for Autonomous Driving, and DrivingGen is a step toward that by first benchmarking open-loop predictive quality and realism. In the current work, all considered generative video world models are designed for open-loop video generation and no standardized closed-loop world generation framework exists yet. Performing a fair, unified closed-loop benchmark is infeasible at this stage. An exciting future direction is to consider closed-loop evaluation for driving world models (\textit{e.g.}, integrating generative models into an interactive simulator like CARLA or combining with closed-loop dataset simulation like Navsim).}

\textcolor{black}{\textbf{Downstream Tasks Metrics and Enriching data modality.} DrivingGen focuses on metrics that directly measure video realism, physical consistency, and controllability in the generated footage itself. One complementary direction is to incorporate metrics from downstream tasks in Autonomous Driving (\textit{e.g.}, how well an autonomous driving stack performs using synthetic videos). However, it may require collecting synchronized multi-camera footage and Map knowledge for a fair and meaningful benchmark. Our current dataset is limited to a single front-view camera feed, which poses challenges for more structural driving generation. A possible future direction is expanding the benchmark to multi-view video and sensor data (LiDAR, HD Map, etc.) to construct a more structured driving world generation and novel metrics (\textit{e.g.}, view consistency) can be proposed.}

\textcolor{black}{\textbf{Evaluation of Scene Controllability and State Transformation.} Evaluating controllability over scene content (\textit{e.g.}, controlling other agents, road layout in the scene) would be highly useful for autonomous applications. We did not include such metrics in our benchmark because implementing a unified evaluation for different models with scene-level control faces challenges both in model support and dataset complexity.
Due to these challenges, we believe it is a great topic for driving world generation which controls scene content and map layout and assessing whether state transformations of the world model are reasonable. One could imagine controlling the presence or behavior of a pedestrian or the configuration of lanes, and checking if the model can follow those constraints.}

\textcolor{black}{\textbf{Counterfactual Reasoning Evaluation.} In our current benchmark, we did not explicitly evaluate counterfactual reasoning. The main reason is that DrivingGen focuses on real driving videos. We are limited to evaluating the scenarios that actually happened. One novel future direction would be counterfactual reasoning evaluation. One can introduce hypothetical events or modifications (like an astronaut on a horse crossing the road, or a car jumping off the ground to overtake other agents, and other unrealistic edge cases) and propose new metrics to check whether the model follows this counterfactual generation.
}

\textcolor{black}{\textbf{Overall Score.} We provide the full table of metrics transparently to evaluate the models, and the average rank serves as a quick summary but not a definitive score. Exploration of a composite, single-index score is an interesting topic, 
which requires normalized distribution and alignment metrics (\textit{e.g.}, FVD and ADE).
}

\bibliography{iclr2026_conference}

\begin{thebibliography}{102}
\providecommand{\natexlab}[1]{#1}
\providecommand{\url}[1]{\texttt{#1}}
\expandafter\ifx\csname urlstyle\endcsname\relax
  \providecommand{\doi}[1]{doi: #1}\else
  \providecommand{\doi}{doi: \begingroup \urlstyle{rm}\Url}\fi

\bibitem[Agarwal et~al.(2025)Agarwal, Ali, Bala, Balaji, Barker, Cai, Chattopadhyay, Chen, Cui, Ding, et~al.]{agarwal2025cosmos}
Niket Agarwal, Arslan Ali, Maciej Bala, Yogesh Balaji, Erik Barker, Tiffany Cai, Prithvijit Chattopadhyay, Yongxin Chen, Yin Cui, Yifan Ding, et~al.
\newblock Cosmos world foundation model platform for physical ai.
\newblock \emph{arXiv preprint arXiv:2501.03575}, 2025.

\bibitem[Alibeigi et~al.(2023)Alibeigi, Ljungbergh, Tonderski, Hess, Lilja, Lindstr{\"o}m, Motorniuk, Fu, Widahl, and Petersson]{zod2023}
Mina Alibeigi, William Ljungbergh, Adam Tonderski, Georg Hess, Adam Lilja, Carl Lindstr{\"o}m, Daria Motorniuk, Junsheng Fu, Jenny Widahl, and Christoffer Petersson.
\newblock Zenseact open dataset: A large-scale and diverse multimodal dataset for autonomous driving.
\newblock In \emph{Proceedings of the IEEE/CVF International Conference on Computer Vision}, pp.\  20178--20188, 2023.

\bibitem[Arai et~al.(2024)Arai, Ishihara, Takahashi, and Yamaguchi]{arai2024actbenchactioncontrollableworld}
Hidehisa Arai, Keishi Ishihara, Tsubasa Takahashi, and Yu~Yamaguchi.
\newblock Act-bench: Towards action controllable world models for autonomous driving, 2024.
\newblock URL \url{https://arxiv.org/abs/2412.05337}.

\bibitem[Arai et~al.(2025)Arai, Miwa, Sasaki, Watanabe, Yamaguchi, Aoki, and Yamamoto]{covla_wacv2025}
Hidehisa Arai, Keita Miwa, Kento Sasaki, Kohei Watanabe, Yu~Yamaguchi, Shunsuke Aoki, and Issei Yamamoto.
\newblock Covla: Comprehensive vision-language-action dataset for autonomous driving.
\newblock In \emph{Proceedings of the Winter Conference on Applications of Computer Vision (WACV)}, pp.\  1933--1943, February 2025.

\bibitem[Bai et~al.(2025)Bai, Chen, Liu, Wang, Ge, Song, Dang, Wang, Wang, Tang, Zhong, Zhu, Yang, Li, Wan, Wang, Ding, Fu, Xu, Ye, Zhang, Xie, Cheng, Zhang, Yang, Xu, and Lin]{qwen25vl2025}
Shuai Bai, Keqin Chen, Xuejing Liu, Jialin Wang, Wenbin Ge, Sibo Song, Kai Dang, Peng Wang, Shijie Wang, Jun Tang, Humen Zhong, Yuanzhi Zhu, Mingkun Yang, Zhaohai Li, Jianqiang Wan, Pengfei Wang, Wei Ding, Zheren Fu, Yiheng Xu, Jiabo Ye, Xi~Zhang, Tianbao Xie, Zesen Cheng, Hang Zhang, Zhibo Yang, Haiyang Xu, and Junyang Lin.
\newblock Qwen2.5-vl technical report.
\newblock \emph{arXiv preprint arXiv:2502.13923}, 2025.

\bibitem[Bansal et~al.(2024)Bansal, Lin, Xie, Zong, Yarom, Bitton, Jiang, Sun, Chang, and Grover]{bansal2024videophyevaluatingphysicalcommonsense}
Hritik Bansal, Zongyu Lin, Tianyi Xie, Zeshun Zong, Michal Yarom, Yonatan Bitton, Chenfanfu Jiang, Yizhou Sun, Kai-Wei Chang, and Aditya Grover.
\newblock Videophy: Evaluating physical commonsense for video generation, 2024.
\newblock URL \url{https://arxiv.org/abs/2406.03520}.

\bibitem[Bartoccioni et~al.(2025)Bartoccioni, Ramzi, Besnier, Venkataramanan, Vu, Xu, Chambon, Gidaris, Odabas, Hurych, Marlet, Boulch, Chen, Éloi Zablocki, Bursuc, Valle, and Cord]{bartoccioni2025vavimvavamautonomousdriving}
Florent Bartoccioni, Elias Ramzi, Victor Besnier, Shashanka Venkataramanan, Tuan-Hung Vu, Yihong Xu, Loick Chambon, Spyros Gidaris, Serkan Odabas, David Hurych, Renaud Marlet, Alexandre Boulch, Mickael Chen, Éloi Zablocki, Andrei Bursuc, Eduardo Valle, and Matthieu Cord.
\newblock Vavim and vavam: Autonomous driving through video generative modeling, 2025.
\newblock URL \url{https://arxiv.org/abs/2502.15672}.

\bibitem[Brooks et~al.(2024)Brooks, Peebles, Holmes, DePue, Guo, Jing, Schnurr, Taylor, Luhman, Luhman, Ng, Wang, and Ramesh]{videoworldsimulators2024}
Tim Brooks, Bill Peebles, Connor Holmes, Will DePue, Yufei Guo, Li~Jing, David Schnurr, Joe Taylor, Troy Luhman, Eric Luhman, Clarence Ng, Ricky Wang, and Aditya Ramesh.
\newblock Video generation models as world simulators.
\newblock \url{https://openai.com/research/video-generation-models-as-world-simulators}, 2024.

\bibitem[Caesar et~al.(2020)Caesar, Bankiti, Lang, Vora, Liong, Xu, Krishnan, Pan, Baldan, and Beijbom]{caesar2020nuscenesmultimodaldatasetautonomous}
Holger Caesar, Varun Bankiti, Alex~H. Lang, Sourabh Vora, Venice~Erin Liong, Qiang Xu, Anush Krishnan, Yu~Pan, Giancarlo Baldan, and Oscar Beijbom.
\newblock nuscenes: A multimodal dataset for autonomous driving, 2020.
\newblock URL \url{https://arxiv.org/abs/1903.11027}.

\bibitem[{CapCut}(2024)]{dreamina2024}
{CapCut}.
\newblock Dreamina.
\newblock \url{https://dreamina.capcut.com/ai-tool/home}, 2024.

\bibitem[Chen et~al.(2025)Chen, Lin, Yang, Lin, Zhu, Fan, Zhang, Chen, Chen, Ma, Xiong, Wang, Pang, Kang, Xu, Jin, Liang, Song, Zhao, Xu, Qiu, Li, Fei, Li, and Zhou]{chen2025skyreelsv2infinitelengthfilmgenerative}
Guibin Chen, Dixuan Lin, Jiangping Yang, Chunze Lin, Junchen Zhu, Mingyuan Fan, Hao Zhang, Sheng Chen, Zheng Chen, Chengcheng Ma, Weiming Xiong, Wei Wang, Nuo Pang, Kang Kang, Zhiheng Xu, Yuzhe Jin, Yupeng Liang, Yubing Song, Peng Zhao, Boyuan Xu, Di~Qiu, Debang Li, Zhengcong Fei, Yang Li, and Yahui Zhou.
\newblock Skyreels-v2: Infinite-length film generative model, 2025.
\newblock URL \url{https://arxiv.org/abs/2504.13074}.

\bibitem[Duan et~al.(2025)Duan, Yu, Chen, Fei-Fei, and Wu]{duan2025worldscoreunifiedevaluationbenchmark}
Haoyi Duan, Hong-Xing Yu, Sirui Chen, Li~Fei-Fei, and Jiajun Wu.
\newblock Worldscore: A unified evaluation benchmark for world generation, 2025.
\newblock URL \url{https://arxiv.org/abs/2504.00983}.

\bibitem[Esser et~al.(2021)Esser, Rombach, and Ommer]{esser2021taming}
Patrick Esser, Robin Rombach, and Bjorn Ommer.
\newblock Taming transformers for high-resolution image synthesis.
\newblock In \emph{Proceedings of the IEEE/CVF conference on computer vision and pattern recognition}, pp.\  12873--12883, 2021.

\bibitem[Fan et~al.(2024)Fan, Luo, Gao, and Zhan]{fan2024aigcbenchcomprehensiveevaluationimagetovideo}
Fanda Fan, Chunjie Luo, Wanling Gao, and Jianfeng Zhan.
\newblock Aigcbench: Comprehensive evaluation of image-to-video content generated by ai, 2024.
\newblock URL \url{https://arxiv.org/abs/2401.01651}.

\bibitem[Fischler \& Bolles(1981)Fischler and Bolles]{Fischler1981-cb}
Martin~A Fischler and Robert~C Bolles.
\newblock Random sample consensus.
\newblock \emph{Commun. ACM}, 24\penalty0 (6):\penalty0 381--395, June 1981.

\bibitem[Gao et~al.(2024)Gao, Yang, Chen, Chitta, Qiu, Geiger, Zhang, and Li]{gao2024vista}
Shenyuan Gao, Jiazhi Yang, Li~Chen, Kashyap Chitta, Yihang Qiu, Andreas Geiger, Jun Zhang, and Hongyang Li.
\newblock Vista: A generalizable driving world model with high fidelity and versatile controllability.
\newblock \emph{Advances in Neural Information Processing Systems}, 37:\penalty0 91560--91596, 2024.

\bibitem[Goodfellow et~al.(2014)Goodfellow, Pouget-Abadie, Mirza, Xu, Warde-Farley, Ozair, Courville, and Bengio]{goodfellow2014generative}
Ian~J Goodfellow, Jean Pouget-Abadie, Mehdi Mirza, Bing Xu, David Warde-Farley, Sherjil Ozair, Aaron Courville, and Yoshua Bengio.
\newblock Generative adversarial nets.
\newblock \emph{Advances in neural information processing systems}, 27, 2014.

\bibitem[HaCohen et~al.(2024)HaCohen, Chiprut, Brazowski, Shalem, Moshe, Richardson, Levin, Shiran, Zabari, Gordon, Panet, Weissbuch, Kulikov, Bitterman, Melumian, and Bibi]{hacohen2024ltxvideo}
Yoav HaCohen, Nisan Chiprut, Benny Brazowski, Daniel Shalem, Dudu Moshe, Eitan Richardson, Eran Levin, Guy Shiran, Nir Zabari, Ori Gordon, Poriya Panet, Sapir Weissbuch, Victor Kulikov, Yaki Bitterman, Zeev Melumian, and Ofir Bibi.
\newblock Ltx-video: Realtime video latent diffusion.
\newblock \emph{arXiv preprint arXiv:2501.00103}, 2024.

\bibitem[Hassan et~al.(2024)Hassan, Stapf, Rahimi, Rezende, Haghighi, Brüggemann, Katircioglu, Zhang, Chen, Saha, Cannici, Aljalbout, Ye, Wang, Davtyan, Salzmann, Scaramuzza, Pollefeys, Favaro, and Alahi]{hassan2024gemgeneralizableegovisionmultimodal}
Mariam Hassan, Sebastian Stapf, Ahmad Rahimi, Pedro M~B Rezende, Yasaman Haghighi, David Brüggemann, Isinsu Katircioglu, Lin Zhang, Xiaoran Chen, Suman Saha, Marco Cannici, Elie Aljalbout, Botao Ye, Xi~Wang, Aram Davtyan, Mathieu Salzmann, Davide Scaramuzza, Marc Pollefeys, Paolo Favaro, and Alexandre Alahi.
\newblock Gem: A generalizable ego-vision multimodal world model for fine-grained ego-motion, object dynamics, and scene composition control, 2024.
\newblock URL \url{https://arxiv.org/abs/2412.11198}.

\bibitem[He et~al.(2025)He, Wang, Wang, Song, Ma, Shao, Liu, and Li]{he2025high}
Dailan He, Xiahong Wang, Shulun Wang, Guanglu Song, Bingqi Ma, Hao Shao, Yu~Liu, and Hongsheng Li.
\newblock High-fidelity diffusion face swapping with id-constrained facial conditioning.
\newblock \emph{arXiv preprint arXiv:2503.22179}, 2025.

\bibitem[Ho \& Salimans(2022)Ho and Salimans]{ho2022classifier}
Jonathan Ho and Tim Salimans.
\newblock Classifier-free diffusion guidance.
\newblock \emph{arXiv preprint arXiv:2207.12598}, 2022.

\bibitem[Ho et~al.(2020)Ho, Jain, and Abbeel]{ho2020denoising}
Jonathan Ho, Ajay Jain, and Pieter Abbeel.
\newblock Denoising diffusion probabilistic models.
\newblock \emph{Advances in neural information processing systems}, 33:\penalty0 6840--6851, 2020.

\bibitem[Hu et~al.(2023)Hu, Russell, Yeo, Murez, Fedoseev, Kendall, Shotton, and Corrado]{hu2023gaia}
Anthony Hu, Lloyd Russell, Hudson Yeo, Zak Murez, George Fedoseev, Alex Kendall, Jamie Shotton, and Gianluca Corrado.
\newblock Gaia-1: A generative world model for autonomous driving.
\newblock \emph{arXiv preprint arXiv:2309.17080}, 2023.

\bibitem[Huang et~al.(2023)Huang, He, Yu, Zhang, Si, Jiang, Zhang, Wu, Jin, Chanpaisit, Wang, Chen, Wang, Lin, Qiao, and Liu]{huang2023vbenchcomprehensivebenchmarksuite}
Ziqi Huang, Yinan He, Jiashuo Yu, Fan Zhang, Chenyang Si, Yuming Jiang, Yuanhan Zhang, Tianxing Wu, Qingyang Jin, Nattapol Chanpaisit, Yaohui Wang, Xinyuan Chen, Limin Wang, Dahua Lin, Yu~Qiao, and Ziwei Liu.
\newblock Vbench: Comprehensive benchmark suite for video generative models, 2023.
\newblock URL \url{https://arxiv.org/abs/2311.17982}.

\bibitem[Huang et~al.(2024)Huang, Zhang, Xu, He, Yu, Dong, Ma, Chanpaisit, Si, Jiang, Wang, Chen, Chen, Wang, Lin, Qiao, and Liu]{huang2024vbenchcomprehensiveversatilebenchmark}
Ziqi Huang, Fan Zhang, Xiaojie Xu, Yinan He, Jiashuo Yu, Ziyue Dong, Qianli Ma, Nattapol Chanpaisit, Chenyang Si, Yuming Jiang, Yaohui Wang, Xinyuan Chen, Ying-Cong Chen, Limin Wang, Dahua Lin, Yu~Qiao, and Ziwei Liu.
\newblock Vbench++: Comprehensive and versatile benchmark suite for video generative models, 2024.
\newblock URL \url{https://arxiv.org/abs/2411.13503}.

\bibitem[{IEEE}(2022)]{ieee_p2020_draft_2022}
{IEEE}.
\newblock Ieee draft standard for automotive system image quality.
\newblock IEEE P2020/D3, 2022.

\bibitem[{IEEE P2020 Working Group}(2018)]{ieee2018ieee}
{IEEE P2020 Working Group}.
\newblock Ieee p2020 automotive imaging white paper.
\newblock \url{https://www.image-engineering.de/content/library/white_paper/P2020_white_paper.pdf}, 2018.

\bibitem[Karnchanachari et~al.(2024)Karnchanachari, Geromichalos, Tan, Li, Eriksen, Yaghoubi, Mehdipour, Bernasconi, Fong, Guo, and Caesar]{karnchanachari2024learningbasedplanningthenuplanbenchmark}
Napat Karnchanachari, Dimitris Geromichalos, Kok~Seang Tan, Nanxiang Li, Christopher Eriksen, Shakiba Yaghoubi, Noushin Mehdipour, Gianmarco Bernasconi, Whye~Kit Fong, Yiluan Guo, and Holger Caesar.
\newblock Towards learning-based planning:the nuplan benchmark for real-world autonomous driving, 2024.
\newblock URL \url{https://arxiv.org/abs/2403.04133}.

\bibitem[Keogh \& Pazzani(2000)Keogh and Pazzani]{Keogh2000ScalingUD}
Eamonn~J. Keogh and Michael~J. Pazzani.
\newblock Scaling up dynamic time warping for datamining applications.
\newblock In \emph{Proceedings of the 6th ACM SIGKDD International Conference on Knowledge Discovery and Data Mining}, pp.\  285--289, 2000.

\bibitem[Kingma \& Welling(2013)Kingma and Welling]{kingma2013auto}
Diederik~P Kingma and Max Welling.
\newblock Auto-encoding variational bayes.
\newblock \emph{arXiv preprint arXiv:1312.6114}, 2013.

\bibitem[Kneip et~al.(2011)Kneip, Scaramuzza, and Siegwart]{5995464}
Laurent Kneip, Davide Scaramuzza, and Roland Siegwart.
\newblock A novel parametrization of the perspective-three-point problem for a direct computation of absolute camera position and orientation.
\newblock In \emph{CVPR 2011}, pp.\  2969--2976, 2011.
\newblock \doi{10.1109/CVPR.2011.5995464}.

\bibitem[Kong et~al.(2024)Kong, Tian, Zhang, Min, Dai, Zhou, Xiong, Li, Wu, Zhang, et~al.]{kong2024hunyuanvideo}
Weijie Kong, Qi~Tian, Zijian Zhang, Rox Min, Zuozhuo Dai, Jin Zhou, Jiangfeng Xiong, Xin Li, Bo~Wu, Jianwei Zhang, et~al.
\newblock Hunyuanvideo: A systematic framework for large video generative models.
\newblock \emph{arXiv preprint arXiv:2412.03603}, 2024.

\bibitem[{Kuaishou}(2024)]{kuaishou2024kling}
{Kuaishou}.
\newblock Kling ai.
\newblock \url{https://klingai.kuaishou.com/}, June 2024.

\bibitem[Kwon et~al.(2025)Kwon, Kim, Go, and Baek]{kwon2025stableworldmodelsmeasuring}
Soonwoo Kwon, Jin-Young Kim, Hyojun Go, and Kyungjune Baek.
\newblock Toward stable world models: Measuring and addressing world instability in generative environments, 2025.
\newblock URL \url{https://arxiv.org/abs/2503.08122}.

\bibitem[Li et~al.(2024)Li, Zhang, Guo, Zhang, Li, Zhang, Zhang, Zhang, Li, Liu, et~al.]{li2024llava}
Bo~Li, Yuanhan Zhang, Dong Guo, Renrui Zhang, Feng Li, Hao Zhang, Kaichen Zhang, Peiyuan Zhang, Yanwei Li, Ziwei Liu, et~al.
\newblock Llava-onevision: Easy visual task transfer.
\newblock \emph{arXiv preprint arXiv:2408.03326}, 2024.

\bibitem[Li et~al.(2025{\natexlab{a}})Li, Fang, Chen, Yang, Cao, Wong, Luo, Wang, Yin, Gonzalez, Stoica, Han, and Lu]{li2025worldmodelbenchjudgingvideogeneration}
Dacheng Li, Yunhao Fang, Yukang Chen, Shuo Yang, Shiyi Cao, Justin Wong, Michael Luo, Xiaolong Wang, Hongxu Yin, Joseph~E. Gonzalez, Ion Stoica, Song Han, and Yao Lu.
\newblock Worldmodelbench: Judging video generation models as world models, 2025{\natexlab{a}}.
\newblock URL \url{https://arxiv.org/abs/2502.20694}.

\bibitem[Li et~al.(2025{\natexlab{b}})Li, Cui, Huang, Ma, Fan, Yang, and Zhong]{li2025mixgrpounlockingflowbasedgrpo}
Junzhe Li, Yutao Cui, Tao Huang, Yinping Ma, Chun Fan, Miles Yang, and Zhao Zhong.
\newblock Mixgrpo: Unlocking flow-based grpo efficiency with mixed ode-sde, 2025{\natexlab{b}}.
\newblock URL \url{https://arxiv.org/abs/2507.21802}.

\bibitem[Li et~al.(2025{\natexlab{c}})Li, Zhou, Guo, Qiu, Xu, Qu, Long, Fan, Li, Fan, Liu, and Yan]{li2025unif2aceunifiedfinegrainedface}
Junzhe Li, Sifan Zhou, Liya Guo, Xuerui Qiu, Linrui Xu, Delin Qu, Tingting Long, Chun Fan, Ming Li, Hehe Fan, Jun Liu, and Shuicheng Yan.
\newblock Unif$^2$ace: A unified fine-grained face understanding and generation model, 2025{\natexlab{c}}.
\newblock URL \url{https://arxiv.org/abs/2503.08120}.

\bibitem[Li et~al.(2025{\natexlab{d}})Li, Wu, Yang, Xu, Liang, Zhang, Wan, and Wang]{li2025driversenavigationworldmodel}
Xiaofan Li, Chenming Wu, Zhao Yang, Zhihao Xu, Dingkang Liang, Yumeng Zhang, Ji~Wan, and Jun Wang.
\newblock Driverse: Navigation world model for driving simulation via multimodal trajectory prompting and motion alignment, 2025{\natexlab{d}}.
\newblock URL \url{https://arxiv.org/abs/2504.18576}.

\bibitem[Liang et~al.(2025)Liang, Zhang, Zhou, Tu, Feng, Li, Zhang, Du, Tan, and Bai]{liang2025seeingfutureperceivingfuture}
Dingkang Liang, Dingyuan Zhang, Xin Zhou, Sifan Tu, Tianrui Feng, Xiaofan Li, Yumeng Zhang, Mingyang Du, Xiao Tan, and Xiang Bai.
\newblock Seeing the future, perceiving the future: A unified driving world model for future generation and perception, 2025.
\newblock URL \url{https://arxiv.org/abs/2503.13587}.

\bibitem[Liao et~al.(2024)Liao, Lu, Zhang, Wan, Wang, Zhao, Zuo, Ye, and Wang]{liao2024evaluationtexttovideogenerationmodels}
Mingxiang Liao, Hannan Lu, Xinyu Zhang, Fang Wan, Tianyu Wang, Yuzhong Zhao, Wangmeng Zuo, Qixiang Ye, and Jingdong Wang.
\newblock Evaluation of text-to-video generation models: A dynamics perspective, 2024.
\newblock URL \url{https://arxiv.org/abs/2407.01094}.

\bibitem[Liu et~al.(2024)Liu, Zhang, Qiu, Huang, Lin, Zhao, Geng, Lin, Jin, Zhang, et~al.]{liu2024sphinx}
Dongyang Liu, Renrui Zhang, Longtian Qiu, Siyuan Huang, Weifeng Lin, Shitian Zhao, Shijie Geng, Ziyi Lin, Peng Jin, Kaipeng Zhang, et~al.
\newblock Sphinx-x: Scaling data and parameters for a family of multi-modal large language models.
\newblock \emph{arXiv preprint arXiv:2402.05935}, 2024.

\bibitem[Lowe(2004)]{Lowe2004-pe}
David~G Lowe.
\newblock Distinctive image features from scale-invariant keypoints.
\newblock \emph{Int. J. Comput. Vis.}, 60\penalty0 (2):\penalty0 91--110, November 2004.

\bibitem[{Luma Labs}(2024)]{luma2024dm}
{Luma Labs}.
\newblock Dream machine.
\newblock \url{https://lumalabs.ai/dream-machine}, June 2024.

\bibitem[{MiniMax}(2024)]{minimax2024hailuo}
{MiniMax}.
\newblock Hailuo ai.
\newblock \url{https://hailuoai.com/video}, September 2024.

\bibitem[Mousakhan et~al.(2025)Mousakhan, Mittal, Galesso, Farid, and Brox]{mousakhan2025orbisovercomingchallengeslonghorizon}
Arian Mousakhan, Sudhanshu Mittal, Silvio Galesso, Karim Farid, and Thomas Brox.
\newblock Orbis: Overcoming challenges of long-horizon prediction in driving world models, 2025.
\newblock URL \url{https://arxiv.org/abs/2507.13162}.

\bibitem[Mur-Artal et~al.(2015)Mur-Artal, Montiel, and Tardós]{7219438}
Raúl Mur-Artal, J.~M.~M. Montiel, and Juan~D. Tardós.
\newblock Orb-slam: A versatile and accurate monocular slam system.
\newblock \emph{IEEE Transactions on Robotics}, 31\penalty0 (5):\penalty0 1147--1163, 2015.
\newblock \doi{10.1109/TRO.2015.2463671}.

\bibitem[Ning et~al.(2023)Ning, Zhu, Xie, Lin, Cui, Yuan, Chen, and Yuan]{ning2023videobenchcomprehensivebenchmarktoolkit}
Munan Ning, Bin Zhu, Yujia Xie, Bin Lin, Jiaxi Cui, Lu~Yuan, Dongdong Chen, and Li~Yuan.
\newblock Video-bench: A comprehensive benchmark and toolkit for evaluating video-based large language models, 2023.
\newblock URL \url{https://arxiv.org/abs/2311.16103}.

\bibitem[{NVIDIA} et~al.(2025){NVIDIA}, Azzolini, Bai, Brandon, Cao, Chattopadhyay, Chen, Chu, Cui, Diamond, Ding, Feng, Ferroni, Govindaraju, Gu, Gururani, Hanafi, Hao, Huffman, Jin, Johnson, Khan, Kurian, Lantz, Lee, Li, Li, Liao, Lin, Lin, Liu, Lu, Luo, Mathau, Ni, Pavao, Ping, Romero, Smelyanskiy, Song, Tchapmi, Wang, Wang, Wang, Wei, Xu, Xu, Yang, Yang, Yang, Zhang, Zeng, and Zhang]{nvidia2025cosmosreason1physicalcommonsense}
{NVIDIA}, Alisson Azzolini, Junjie Bai, Hannah Brandon, Jiaxin Cao, Prithvijit Chattopadhyay, Huayu Chen, Jinju Chu, Yin Cui, Jenna Diamond, Yifan Ding, Liang Feng, Francesco Ferroni, Rama Govindaraju, Jinwei Gu, Siddharth Gururani, Imad~El Hanafi, Zekun Hao, Jacob Huffman, Jingyi Jin, Brendan Johnson, Rizwan Khan, George Kurian, Elena Lantz, Nayeon Lee, Zhaoshuo Li, Xuan Li, Maosheng Liao, Tsung-Yi Lin, Yen-Chen Lin, Ming-Yu Liu, Xiangyu Lu, Alice Luo, Andrew Mathau, Yun Ni, Lindsey Pavao, Wei Ping, David~W. Romero, Misha Smelyanskiy, Shuran Song, Lyne Tchapmi, Andrew~Z. Wang, Boxin Wang, Haoxiang Wang, Fangyin Wei, Jiashu Xu, Yao Xu, Dinghao Yang, Xiaodong Yang, Zhuolin Yang, Jingxu Zhang, Xiaohui Zeng, and Zhe Zhang.
\newblock Cosmos-reason1: From physical common sense to embodied reasoning, 2025.

\bibitem[{NVIDIA Cosmos}(2025)]{cosmospredict2_2025}
{NVIDIA Cosmos}.
\newblock Cosmos predict-2.
\newblock \url{https://github.com/nvidia-cosmos/cosmos-predict2}, 2025.

\bibitem[Peebles \& Xie(2023)Peebles and Xie]{peebles2023scalable}
William Peebles and Saining Xie.
\newblock Scalable diffusion models with transformers.
\newblock In \emph{Proceedings of the IEEE/CVF international conference on computer vision}, pp.\  4195--4205, 2023.

\bibitem[Piccinelli et~al.(2025)Piccinelli, Sakaridis, Yang, Segu, Li, Abbeloos, and Gool]{piccinelli2025unidepthv2}
Luigi Piccinelli, Christos Sakaridis, Yung-Hsu Yang, Mattia Segu, Siyuan Li, Wim Abbeloos, and Luc~Van Gool.
\newblock {U}ni{D}epth{V2}: Universal monocular metric depth estimation made simpler, 2025.
\newblock URL \url{https://arxiv.org/abs/2502.20110}.

\bibitem[{Pika Labs}(2024)]{pika2024pika}
{Pika Labs}.
\newblock Pika 1.5.
\newblock \url{https://pika.art/}, October 2024.

\bibitem[{PixVerse}(2023)]{pixverse2023}
{PixVerse}.
\newblock Pixverse.
\newblock \url{https://pixverse.ai/}, 2023.

\bibitem[Qin et~al.(2024)Qin, Shi, Yu, Wang, Zhou, Li, Yin, Liu, Sheng, Shao, Bai, Ouyang, and Zhang]{qin2024worldsimbenchvideogenerationmodels}
Yiran Qin, Zhelun Shi, Jiwen Yu, Xijun Wang, Enshen Zhou, Lijun Li, Zhenfei Yin, Xihui Liu, Lu~Sheng, Jing Shao, Lei Bai, Wanli Ouyang, and Ruimao Zhang.
\newblock Worldsimbench: Towards video generation models as world simulators, 2024.
\newblock URL \url{https://arxiv.org/abs/2410.18072}.

\bibitem[Qu et~al.(2024)Qu, Yan, Wang, Yin, Chen, Xu, Zhang, Zhao, and Li]{qu2024implicit}
Delin Qu, Chi Yan, Dong Wang, Jie Yin, Qizhi Chen, Dan Xu, Yiting Zhang, Bin Zhao, and Xuelong Li.
\newblock Implicit event-rgbd neural slam.
\newblock In \emph{Proceedings of the IEEE/CVF conference on computer vision and pattern recognition}, pp.\  19584--19594, 2024.

\bibitem[Qu et~al.(2025{\natexlab{a}})Qu, Song, Chen, Chen, Gao, Ye, Lv, Shi, Ren, Ruan, et~al.]{qu2025eo}
Delin Qu, Haoming Song, Qizhi Chen, Zhaoqing Chen, Xianqiang Gao, Xinyi Ye, Qi~Lv, Modi Shi, Guanghui Ren, Cheng Ruan, et~al.
\newblock Eo-1: Interleaved vision-text-action pretraining for general robot control.
\newblock \emph{arXiv preprint arXiv:2508.21112}, 2025{\natexlab{a}}.

\bibitem[Qu et~al.(2025{\natexlab{b}})Qu, Song, Chen, Yao, Ye, Ding, Wang, Gu, Zhao, Wang, et~al.]{qu2025spatialvla}
Delin Qu, Haoming Song, Qizhi Chen, Yuanqi Yao, Xinyi Ye, Yan Ding, Zhigang Wang, JiaYuan Gu, Bin Zhao, Dong Wang, et~al.
\newblock Spatialvla: Exploring spatial representations for visual-language-action model.
\newblock \emph{arXiv preprint arXiv:2501.15830}, 2025{\natexlab{b}}.

\bibitem[Ravi et~al.(2024)Ravi, Gabeur, Hu, Hu, Ryali, Ma, Khedr, R{\"a}dle, Rolland, Gustafson, Mintun, Pan, Alwala, Carion, Wu, Girshick, Doll{\'a}r, and Feichtenhofer]{ravi2024sam2}
Nikhila Ravi, Valentin Gabeur, Yuan-Ting Hu, Ronghang Hu, Chaitanya Ryali, Tengyu Ma, Haitham Khedr, Roman R{\"a}dle, Chloe Rolland, Laura Gustafson, Eric Mintun, Junting Pan, Kalyan~Vasudev Alwala, Nicolas Carion, Chao-Yuan Wu, Ross Girshick, Piotr Doll{\'a}r, and Christoph Feichtenhofer.
\newblock Sam 2: Segment anything in images and videos.
\newblock \emph{arXiv preprint arXiv:2408.00714}, 2024.
\newblock URL \url{https://arxiv.org/abs/2408.00714}.

\bibitem[Rombach et~al.(2022)Rombach, Blattmann, Lorenz, Esser, and Ommer]{rombach2022high}
Robin Rombach, Andreas Blattmann, Dominik Lorenz, Patrick Esser, and Bj{\"o}rn Ommer.
\newblock High-resolution image synthesis with latent diffusion models.
\newblock In \emph{Proceedings of the IEEE/CVF conference on computer vision and pattern recognition}, pp.\  10684--10695, 2022.

\bibitem[{Runway}(2024)]{runway2024gen3}
{Runway}.
\newblock Gen-3.
\newblock \url{https://runwayml.com/}, June 2024.

\bibitem[Russell et~al.(2025)Russell, Hu, Bertoni, Fedoseev, Shotton, Arani, and Corrado]{russell2025gaia2controllablemultiviewgenerative}
Lloyd Russell, Anthony Hu, Lorenzo Bertoni, George Fedoseev, Jamie Shotton, Elahe Arani, and Gianluca Corrado.
\newblock Gaia-2: A controllable multi-view generative world model for autonomous driving, 2025.
\newblock URL \url{https://arxiv.org/abs/2503.20523}.

\bibitem[Sch\"{o}nberger \& Frahm(2016)Sch\"{o}nberger and Frahm]{schoenberger2016sfm}
Johannes~Lutz Sch\"{o}nberger and Jan-Michael Frahm.
\newblock Structure-from-motion revisited.
\newblock In \emph{Conference on Computer Vision and Pattern Recognition (CVPR)}, 2016.

\bibitem[Sch\"{o}nberger et~al.(2016)Sch\"{o}nberger, Zheng, Pollefeys, and Frahm]{schoenberger2016mvs}
Johannes~Lutz Sch\"{o}nberger, Enliang Zheng, Marc Pollefeys, and Jan-Michael Frahm.
\newblock Pixelwise view selection for unstructured multi-view stereo.
\newblock In \emph{European Conference on Computer Vision (ECCV)}, 2016.

\bibitem[Shao et~al.(2023{\natexlab{a}})Shao, Wang, Chen, Li, and Liu]{shao2023safety}
Hao Shao, Letian Wang, Ruobing Chen, Hongsheng Li, and Yu~Liu.
\newblock Safety-enhanced autonomous driving using interpretable sensor fusion transformer.
\newblock In \emph{Conference on Robot Learning}, pp.\  726--737. PMLR, 2023{\natexlab{a}}.

\bibitem[Shao et~al.(2023{\natexlab{b}})Shao, Wang, Chen, Waslander, Li, and Liu]{shao2023reasonnet}
Hao Shao, Letian Wang, Ruobing Chen, Steven~L Waslander, Hongsheng Li, and Yu~Liu.
\newblock Reasonnet: End-to-end driving with temporal and global reasoning.
\newblock In \emph{Proceedings of the IEEE/CVF conference on computer vision and pattern recognition}, pp.\  13723--13733, 2023{\natexlab{b}}.

\bibitem[Shao et~al.(2024{\natexlab{a}})Shao, Hu, Wang, Song, Waslander, Liu, and Li]{shao2024lmdrive}
Hao Shao, Yuxuan Hu, Letian Wang, Guanglu Song, Steven~L Waslander, Yu~Liu, and Hongsheng Li.
\newblock Lmdrive: Closed-loop end-to-end driving with large language models.
\newblock In \emph{Proceedings of the IEEE/CVF Conference on Computer Vision and Pattern Recognition}, pp.\  15120--15130, 2024{\natexlab{a}}.

\bibitem[Shao et~al.(2024{\natexlab{b}})Shao, Qian, Xiao, Song, Zong, Wang, Liu, and Li]{shao2024visual}
Hao Shao, Shengju Qian, Han Xiao, Guanglu Song, Zhuofan Zong, Letian Wang, Yu~Liu, and Hongsheng Li.
\newblock Visual cot: Advancing multi-modal language models with a comprehensive dataset and benchmark for chain-of-thought reasoning.
\newblock \emph{Advances in Neural Information Processing Systems}, 37:\penalty0 8612--8642, 2024{\natexlab{b}}.

\bibitem[Shao et~al.(2024{\natexlab{c}})Shao, Wang, Zhou, Song, He, Qin, Zong, Ma, Liu, and Li]{shao2024vividface}
Hao Shao, Shulun Wang, Yang Zhou, Guanglu Song, Dailan He, Shuo Qin, Zhuofan Zong, Bingqi Ma, Yu~Liu, and Hongsheng Li.
\newblock Vividface: A diffusion-based hybrid framework for high-fidelity video face swapping.
\newblock \emph{arXiv preprint arXiv:2412.11279}, 2024{\natexlab{c}}.

\bibitem[Shi et~al.(2023)Shi, Jiang, Dai, and Schiele]{shi2023motiontransformerglobalintention}
Shaoshuai Shi, Li~Jiang, Dengxin Dai, and Bernt Schiele.
\newblock Motion transformer with global intention localization and local movement refinement, 2023.
\newblock URL \url{https://arxiv.org/abs/2209.13508}.

\bibitem[Sim{\'e}oni et~al.(2025)Sim{\'e}oni, Vo, Seitzer, Baldassarre, Oquab, Jose, Khalidov, Szafraniec, Yi, Ramamonjisoa, Massa, Haziza, Wehrstedt, Wang, Darcet, Moutakanni, Sentana, Roberts, Vedaldi, Tolan, Brandt, Couprie, Mairal, J{\'e}gou, Labatut, and Bojanowski]{simeoni2025dinov3}
Oriane Sim{\'e}oni, Huy~V. Vo, Maximilian Seitzer, Federico Baldassarre, Maxime Oquab, Cijo Jose, Vasil Khalidov, Marc Szafraniec, Seungeun Yi, Micha{\"e}l Ramamonjisoa, Francisco Massa, Daniel Haziza, Luca Wehrstedt, Jianyuan Wang, Timoth{\'e}e Darcet, Th{\'e}o Moutakanni, Leonel Sentana, Claire Roberts, Andrea Vedaldi, Jamie Tolan, John Brandt, Camille Couprie, Julien Mairal, Herv{\'e} J{\'e}gou, Patrick Labatut, and Piotr Bojanowski.
\newblock Dinov3, 2025.

\bibitem[Skorokhodov et~al.(2022)Skorokhodov, Tulyakov, and Elhoseiny]{skorokhodov2022styleganvcontinuousvideogenerator}
Ivan Skorokhodov, Sergey Tulyakov, and Mohamed Elhoseiny.
\newblock Stylegan-v: A continuous video generator with the price, image quality and perks of stylegan2, 2022.
\newblock URL \url{https://arxiv.org/abs/2112.14683}.

\bibitem[Sun et~al.(2020)Sun, Kretzschmar, Dotiwalla, Chouard, Patnaik, Tsui, Guo, Zhou, Chai, Caine, Vasudevan, Han, Ngiam, Zhao, Timofeev, Ettinger, Krivokon, Gao, Joshi, Zhao, Cheng, Zhang, Shlens, Chen, and Anguelov]{sun2020scalabilityperceptionautonomousdriving}
Pei Sun, Henrik Kretzschmar, Xerxes Dotiwalla, Aurelien Chouard, Vijaysai Patnaik, Paul Tsui, James Guo, Yin Zhou, Yuning Chai, Benjamin Caine, Vijay Vasudevan, Wei Han, Jiquan Ngiam, Hang Zhao, Aleksei Timofeev, Scott Ettinger, Maxim Krivokon, Amy Gao, Aditya Joshi, Sheng Zhao, Shuyang Cheng, Yu~Zhang, Jonathon Shlens, Zhifeng Chen, and Dragomir Anguelov.
\newblock Scalability in perception for autonomous driving: Waymo open dataset, 2020.
\newblock URL \url{https://arxiv.org/abs/1912.04838}.

\bibitem[Teed \& Deng(2022)Teed and Deng]{teed2022droidslamdeepvisualslam}
Zachary Teed and Jia Deng.
\newblock Droid-slam: Deep visual slam for monocular, stereo, and rgb-d cameras, 2022.
\newblock URL \url{https://arxiv.org/abs/2108.10869}.

\bibitem[Unterthiner et~al.(2019)Unterthiner, van Steenkiste, Kurach, Marinier, Michalski, and Gelly]{unterthiner2019accurategenerativemodelsvideo}
Thomas Unterthiner, Sjoerd van Steenkiste, Karol Kurach, Raphael Marinier, Marcin Michalski, and Sylvain Gelly.
\newblock Towards accurate generative models of video: A new metric \& challenges, 2019.
\newblock URL \url{https://arxiv.org/abs/1812.01717}.

\bibitem[Wan et~al.(2025)Wan, Wang, Ai, Wen, Mao, Xie, Chen, Yu, Zhao, Yang, Zeng, Wang, Zhang, Zhou, Wang, Chen, Zhu, Zhao, Yan, Huang, Feng, Zhang, Li, Wu, Chu, Feng, Zhang, Sun, Fang, Wang, Gui, Weng, Shen, Lin, Wang, Wang, Zhou, Wang, Shen, Yu, Shi, Huang, Xu, Kou, Lv, Li, Liu, Wang, Zhang, Huang, Li, Wu, Liu, Pan, Zheng, Hong, Shi, Feng, Jiang, Han, Wu, and Liu]{wan2025wanopenadvancedlargescale}
Team Wan, Ang Wang, Baole Ai, Bin Wen, Chaojie Mao, Chen-Wei Xie, Di~Chen, Feiwu Yu, Haiming Zhao, Jianxiao Yang, Jianyuan Zeng, Jiayu Wang, Jingfeng Zhang, Jingren Zhou, Jinkai Wang, Jixuan Chen, Kai Zhu, Kang Zhao, Keyu Yan, Lianghua Huang, Mengyang Feng, Ningyi Zhang, Pandeng Li, Pingyu Wu, Ruihang Chu, Ruili Feng, Shiwei Zhang, Siyang Sun, Tao Fang, Tianxing Wang, Tianyi Gui, Tingyu Weng, Tong Shen, Wei Lin, Wei Wang, Wei Wang, Wenmeng Zhou, Wente Wang, Wenting Shen, Wenyuan Yu, Xianzhong Shi, Xiaoming Huang, Xin Xu, Yan Kou, Yangyu Lv, Yifei Li, Yijing Liu, Yiming Wang, Yingya Zhang, Yitong Huang, Yong Li, You Wu, Yu~Liu, Yulin Pan, Yun Zheng, Yuntao Hong, Yupeng Shi, Yutong Feng, Zeyinzi Jiang, Zhen Han, Zhi-Fan Wu, and Ziyu Liu.
\newblock Wan: Open and advanced large-scale video generative models, 2025.
\newblock URL \url{https://arxiv.org/abs/2503.20314}.

\bibitem[Wang et~al.(2024{\natexlab{a}})Wang, Chen, Liu, Chen, Lin, Han, and Ding]{wang2024yolov10}
Ao~Wang, Hui Chen, Lihao Liu, Kai Chen, Zijia Lin, Jungong Han, and Guiguang Ding.
\newblock Yolov10: Real-time end-to-end object detection.
\newblock \emph{arXiv preprint arXiv:2405.14458}, 2024{\natexlab{a}}.

\bibitem[Wang et~al.(2022)Wang, Chan, and Loy]{wang2022exploringclipassessinglook}
Jianyi Wang, Kelvin C.~K. Chan, and Chen~Change Loy.
\newblock Exploring clip for assessing the look and feel of images, 2022.
\newblock URL \url{https://arxiv.org/abs/2207.12396}.

\bibitem[Wang et~al.(2021)Wang, Sun, Tomizuka, and Zhan]{wang2021socially}
Letian Wang, Liting Sun, Masayoshi Tomizuka, and Wei Zhan.
\newblock Socially-compatible behavior design of autonomous vehicles with verification on real human data.
\newblock \emph{IEEE Robotics and Automation Letters}, 6\penalty0 (2):\penalty0 3421--3428, 2021.

\bibitem[Wang et~al.(2023)Wang, Liu, Shao, Wang, Chen, Liu, and Waslander]{wang2023efficient}
Letian Wang, Jie Liu, Hao Shao, Wenshuo Wang, Ruobing Chen, Yu~Liu, and Steven~L Waslander.
\newblock Efficient reinforcement learning for autonomous driving with parameterized skills and priors.
\newblock \emph{arXiv preprint arXiv:2305.04412}, 2023.

\bibitem[Wang et~al.(2024{\natexlab{b}})Wang, Kim, Yang, Yu, Ivanovic, Waslander, Wang, Fidler, Pavone, and Karkus]{wang2024distillnerf}
Letian Wang, Seung~Wook Kim, Jiawei Yang, Cunjun Yu, Boris Ivanovic, Steven Waslander, Yue Wang, Sanja Fidler, Marco Pavone, and Peter Karkus.
\newblock Distillnerf: Perceiving 3d scenes from single-glance images by distilling neural fields and foundation model features.
\newblock \emph{Advances in Neural Information Processing Systems}, 37:\penalty0 62334--62361, 2024{\natexlab{b}}.

\bibitem[Wang et~al.(2025)Wang, Lavoie, Papais, Nisar, Chen, Ding, Ivanovic, Shao, Abuduweili, Cook, et~al.]{wang2025deployable}
Letian Wang, Marc-Antoine Lavoie, Sandro Papais, Barza Nisar, Yuxiao Chen, Wenhao Ding, Boris Ivanovic, Hao Shao, Abulikemu Abuduweili, Evan Cook, et~al.
\newblock Deployable and generalizable motion prediction: Taxonomy, open challenges and future directions.
\newblock \emph{arXiv preprint arXiv:2505.09074}, 2025.

\bibitem[Wang et~al.(2024{\natexlab{c}})Wang, Zhu, Huang, Chen, Zhu, and Lu]{wang2024drivedreamer}
Xiaofeng Wang, Zheng Zhu, Guan Huang, Xinze Chen, Jiagang Zhu, and Jiwen Lu.
\newblock Drivedreamer: Towards real-world-drive world models for autonomous driving.
\newblock In \emph{European conference on computer vision}, pp.\  55--72. Springer, 2024{\natexlab{c}}.

\bibitem[Wang et~al.(2024{\natexlab{d}})Wang, Lipson, and Deng]{wang2024searaftsimpleefficientaccurate}
Yihan Wang, Lahav Lipson, and Jia Deng.
\newblock Sea-raft: Simple, efficient, accurate raft for optical flow, 2024{\natexlab{d}}.
\newblock URL \url{https://arxiv.org/abs/2405.14793}.

\bibitem[Wang et~al.(2024{\natexlab{e}})Wang, Cheng, He, Wang, Dai, Chen, Xia, and Zhang]{wang2024drivingdojodatasetadvancinginteractive}
Yuqi Wang, Ke~Cheng, Jiawei He, Qitai Wang, Hengchen Dai, Yuntao Chen, Fei Xia, and Zhaoxiang Zhang.
\newblock Drivingdojo dataset: Advancing interactive and knowledge-enriched driving world model, 2024{\natexlab{e}}.
\newblock URL \url{https://arxiv.org/abs/2410.10738}.

\bibitem[Wang et~al.(2024{\natexlab{f}})Wang, He, Fan, Li, Chen, and Zhang]{wang2024driving}
Yuqi Wang, Jiawei He, Lue Fan, Hongxin Li, Yuntao Chen, and Zhaoxiang Zhang.
\newblock Driving into the future: Multiview visual forecasting and planning with world model for autonomous driving.
\newblock In \emph{Proceedings of the IEEE/CVF Conference on Computer Vision and Pattern Recognition}, pp.\  14749--14759, 2024{\natexlab{f}}.

\bibitem[Wang et~al.(2024{\natexlab{g}})Wang, Li, Hao, Hu, and Song]{wang2024mattersnovelvisualphysicsbased}
Zihan Wang, Songlin Li, Lingyan Hao, Xinyu Hu, and Bowen Song.
\newblock What you see is what matters: A novel visual and physics-based metric for evaluating video generation quality, 2024{\natexlab{g}}.
\newblock URL \url{https://arxiv.org/abs/2411.13609}.

\bibitem[Yang et~al.(2024{\natexlab{a}})Yang, Huang, Chai, Jiang, and Hwang]{yang2024samuraiadaptingsegmentmodel}
Cheng-Yen Yang, Hsiang-Wei Huang, Wenhao Chai, Zhongyu Jiang, and Jenq-Neng Hwang.
\newblock Samurai: Adapting segment anything model for zero-shot visual tracking with motion-aware memory, 2024{\natexlab{a}}.
\newblock URL \url{https://arxiv.org/abs/2411.11922}.

\bibitem[Yang et~al.(2024{\natexlab{b}})Yang, Gao, Qiu, Chen, Li, Dai, Chitta, Wu, Zeng, Luo, et~al.]{yang2024generalized}
Jiazhi Yang, Shenyuan Gao, Yihang Qiu, Li~Chen, Tianyu Li, Bo~Dai, Kashyap Chitta, Penghao Wu, Jia Zeng, Ping Luo, et~al.
\newblock Generalized predictive model for autonomous driving.
\newblock In \emph{Proceedings of the IEEE/CVF Conference on Computer Vision and Pattern Recognition}, pp.\  14662--14672, 2024{\natexlab{b}}.

\bibitem[Yang et~al.(2025)Yang, Chitta, Gao, Chen, Shao, Jia, Li, Geiger, Yue, and Chen]{yang2025resimreliableworldsimulation}
Jiazhi Yang, Kashyap Chitta, Shenyuan Gao, Long Chen, Yuqian Shao, Xiaosong Jia, Hongyang Li, Andreas Geiger, Xiangyu Yue, and Li~Chen.
\newblock Resim: Reliable world simulation for autonomous driving, 2025.
\newblock URL \url{https://arxiv.org/abs/2506.09981}.

\bibitem[Yang et~al.(2024{\natexlab{c}})Yang, Kang, Huang, Zhao, Xu, Feng, and Zhao]{yang2024depthv2}
Lihe Yang, Bingyi Kang, Zilong Huang, Zhen Zhao, Xiaogang Xu, Jiashi Feng, and Hengshuang Zhao.
\newblock Depth anything v2, 2024{\natexlab{c}}.
\newblock URL \url{https://arxiv.org/abs/2406.09414}.

\bibitem[Yang et~al.(2024{\natexlab{d}})Yang, Chen, Sun, and Li]{yang2024visual}
Zetong Yang, Li~Chen, Yanan Sun, and Hongyang Li.
\newblock Visual point cloud forecasting enables scalable autonomous driving.
\newblock In \emph{Proceedings of the IEEE/CVF Conference on Computer Vision and Pattern Recognition}, pp.\  14673--14684, 2024{\natexlab{d}}.

\bibitem[Yang et~al.(2024{\natexlab{e}})Yang, Teng, Zheng, Ding, Huang, Xu, Yang, Hong, Zhang, Feng, et~al.]{yang2024cogvideox}
Zhuoyi Yang, Jiayan Teng, Wendi Zheng, Ming Ding, Shiyu Huang, Jiazheng Xu, Yuanming Yang, Wenyi Hong, Xiaohan Zhang, Guanyu Feng, et~al.
\newblock Cogvideox: Text-to-video diffusion models with an expert transformer.
\newblock \emph{arXiv preprint arXiv:2408.06072}, 2024{\natexlab{e}}.

\bibitem[Yue et~al.(2025)Yue, Huang, Liao, Chen, Zhou, Chen, Yao, and Ren]{yue2025ewmbenchevaluatingscenemotion}
Hu~Yue, Siyuan Huang, Yue Liao, Shengcong Chen, Pengfei Zhou, Liliang Chen, Maoqing Yao, and Guanghui Ren.
\newblock Ewmbench: Evaluating scene, motion, and semantic quality in embodied world models, 2025.
\newblock URL \url{https://arxiv.org/abs/2505.09694}.

\bibitem[Zhang et~al.(2025)Zhang, Guo, Zhao, Fu, Duan, Hu, Chng, Wang, Chen, Xu, Luo, and Zhang]{zhang2025unifiedmultimodalunderstandinggeneration}
Xinjie Zhang, Jintao Guo, Shanshan Zhao, Minghao Fu, Lunhao Duan, Jiakui Hu, Yong~Xien Chng, Guo-Hua Wang, Qing-Guo Chen, Zhao Xu, Weihua Luo, and Kaifu Zhang.
\newblock Unified multimodal understanding and generation models: Advances, challenges, and opportunities, 2025.
\newblock URL \url{https://arxiv.org/abs/2505.02567}.

\bibitem[Zhao et~al.(2024)Zhao, Wang, Zhu, Chen, Huang, Bao, and Wang]{zhao2025drivedreamer}
Guosheng Zhao, Xiaofeng Wang, Zheng Zhu, Xinze Chen, Guan Huang, Xiaoyi Bao, and Xingang Wang.
\newblock Drivedreamer-2: Llm-enhanced world models for diverse driving video generation, 2024.
\newblock URL \url{https://arxiv.org/abs/2403.06845}.

\bibitem[Zheng et~al.(2024{\natexlab{a}})Zheng, Chen, Huang, Zhang, Duan, and Lu]{zheng2024occworld}
Wenzhao Zheng, Weiliang Chen, Yuanhui Huang, Borui Zhang, Yueqi Duan, and Jiwen Lu.
\newblock Occworld: Learning a 3d occupancy world model for autonomous driving.
\newblock In \emph{European conference on computer vision}, pp.\  55--72. Springer, 2024{\natexlab{a}}.

\bibitem[Zheng et~al.(2024{\natexlab{b}})Zheng, Song, Guo, Zhang, and Chen]{zheng2024genadgenerativeendtoendautonomous}
Wenzhao Zheng, Ruiqi Song, Xianda Guo, Chenming Zhang, and Long Chen.
\newblock Genad: Generative end-to-end autonomous driving, 2024{\natexlab{b}}.
\newblock URL \url{https://arxiv.org/abs/2402.11502}.

\bibitem[Zheng et~al.(2024{\natexlab{c}})Zheng, Xia, Huang, Zuo, Zhou, and Lu]{zheng2024doe1closedloopautonomousdriving}
Wenzhao Zheng, Zetian Xia, Yuanhui Huang, Sicheng Zuo, Jie Zhou, and Jiwen Lu.
\newblock Doe-1: Closed-loop autonomous driving with large world model, 2024{\natexlab{c}}.
\newblock URL \url{https://arxiv.org/abs/2412.09627}.

\bibitem[Zhou et~al.(2025)Zhou, Shao, Wang, Waslander, Li, and Liu]{zhou2025smartpretrain}
Yang Zhou, Hao Shao, Letian Wang, Steven~L. Waslander, Hongsheng Li, and Yu~Liu.
\newblock Smartpretrain: Model-agnostic and dataset-agnostic representation learning for motion prediction.
\newblock In \emph{The Thirteenth International Conference on Learning Representations}, 2025.
\newblock URL \url{https://openreview.net/forum?id=Bmzv2Gch9v}.

\bibitem[Zong et~al.(2024{\natexlab{a}})Zong, Jiang, Ma, Song, Shao, Shen, Liu, and Li]{zong2024easyref}
Zhuofan Zong, Dongzhi Jiang, Bingqi Ma, Guanglu Song, Hao Shao, Dazhong Shen, Yu~Liu, and Hongsheng Li.
\newblock Easyref: Omni-generalized group image reference for diffusion models via multimodal llm.
\newblock In \emph{Forty-second International Conference on Machine Learning}, 2024{\natexlab{a}}.

\bibitem[Zong et~al.(2024{\natexlab{b}})Zong, Ma, Shen, Song, Shao, Jiang, Li, and Liu]{zong2024mova}
Zhuofan Zong, Bingqi Ma, Dazhong Shen, Guanglu Song, Hao Shao, Dongzhi Jiang, Hongsheng Li, and Yu~Liu.
\newblock Mova: Adapting mixture of vision experts to multimodal context.
\newblock \emph{Advances in Neural Information Processing Systems}, 37:\penalty0 103305--103333, 2024{\natexlab{b}}.

\end{thebibliography}
\bibliographystyle{iclr2026_conference}

\newpage
\appendix

\section{Related Works}
\label{app:related_works}

\subsection{Generative World Models and their application in Driving}
Driven by advances in image generative modeling~\citep{kingma2013auto,goodfellow2014generative,esser2021taming,ho2020denoising, peebles2023scalable,zong2024easyref, he2025high}, the landscape of large-scale video models has evolved
significantly, particularly in diffusion-based frameworks. Closed-source models~\citep{videoworldsimulators2024,kuaishou2024kling,luma2024dm,runway2024gen3,pixverse2023,dreamina2024,minimax2024hailuo,pika2024pika,shao2024vividface}, mainly developed by major technology companies, aim at high-quality, professional video generation with extensive resources invested. Sora~\citep{videoworldsimulators2024}, introduced by OpenAI, marked a significant leap in Video Generation. Open-source models~\citep{rombach2022high,ho2022classifier,hacohen2024ltxvideo,kong2024hunyuanvideo,wan2025wanopenadvancedlargescale,yang2024cogvideox,agarwal2025cosmos}, typically based on stable diffusion~\citep{rombach2022high} and flow matching~\citep{li2025mixgrpounlockingflowbasedgrpo}, are quickly expanding and making real contributions to video generation as well. Wan~\citep{wan2025wanopenadvancedlargescale}, an open-source model, is widely used for video generation and has achieved SOTA results on many benchmarks. Recent years have also seen remarkable progress in both multimodal understanding and generation models~\citep{li2025unif2aceunifiedfinegrainedface, zhang2025unifiedmultimodalunderstandinggeneration}. 

Besides general video generation, driving-focused generative models use sensor data such as lidar point clouds~\citep{zheng2024occworld,yang2024visual} or images~\citep{gao2024vista,hassan2024gemgeneralizableegovisionmultimodal,hu2023gaia,wang2024drivedreamer,wang2024driving,yang2024generalized,zhao2025drivedreamer}. 
Since this work emphasizes video generation, we focus on image-based methods.  
Early approaches before Vista~\citep{gao2024vista} rely on multi-view RGB inputs and high-definition maps or 3D boxes, limiting generalization to new datasets and open-domain videos.  
Vista-based methods~\citep{hassan2024gemgeneralizableegovisionmultimodal,li2025driversenavigationworldmodel,mousakhan2025orbisovercomingchallengeslonghorizon} simplify inputs to a single front-view image with optional ego trajectories, improving scalability to YouTube videos and enabling broader open-domain evaluation.

\subsection{Benchmarks for evaluating generative world models}

The rapid progress of open- and closed-source video generation has driven the creation of many benchmarks~\citep{huang2023vbenchcomprehensivebenchmarksuite,huang2024vbenchcomprehensiveversatilebenchmark,bansal2024videophyevaluatingphysicalcommonsense,ning2023videobenchcomprehensivebenchmarktoolkit,liao2024evaluationtexttovideogenerationmodels,fan2024aigcbenchcomprehensiveevaluationimagetovideo,wang2024mattersnovelvisualphysicsbased}, such as VBench, which evaluates models with multifaceted metrics based on human-collected prompts.  
Recently, evaluations have expanded to open, dynamic, and complex world-simulation scenarios~\citep{yue2025ewmbenchevaluatingscenemotion,duan2025worldscoreunifiedevaluationbenchmark,li2025driversenavigationworldmodel,qin2024worldsimbenchvideogenerationmodels,kwon2025stableworldmodelsmeasuring}.  
WorldScore~\citep{duan2025worldscoreunifiedevaluationbenchmark} measures generated videos using explicit camera trajectory layouts.  
However, a comprehensive driving-world benchmark is still lacking due to limited test sample diversity, heterogeneous input modalities, and the absence of driving-specific metrics.  
Recent works~\citep{gao2024vista,hassan2024gemgeneralizableegovisionmultimodal} mainly adopt Frechet Video Distance (FVD) and Average Displacement Error (ADE) for trajectory alignment, while GEM~\citep{hassan2024gemgeneralizableegovisionmultimodal} adds human video evaluations that are subjective and hard to scale.  
The closest effort, ACT-Bench~\citep{arai2024actbenchactioncontrollableworld}, focuses solely on trajectory alignment and overlooks key aspects such as video and trajectory distribution, quality, and temporal consistency.

\newpage
\section{Appendix}

\subsection{Gallery of the Ego-conditioned Track}
\label{app:align}

\begin{figure*}
  \centering

  \begin{subfigure}[t]{0.5\linewidth}\vspace{0pt}
    \centering
    \includegraphics[width=\linewidth]{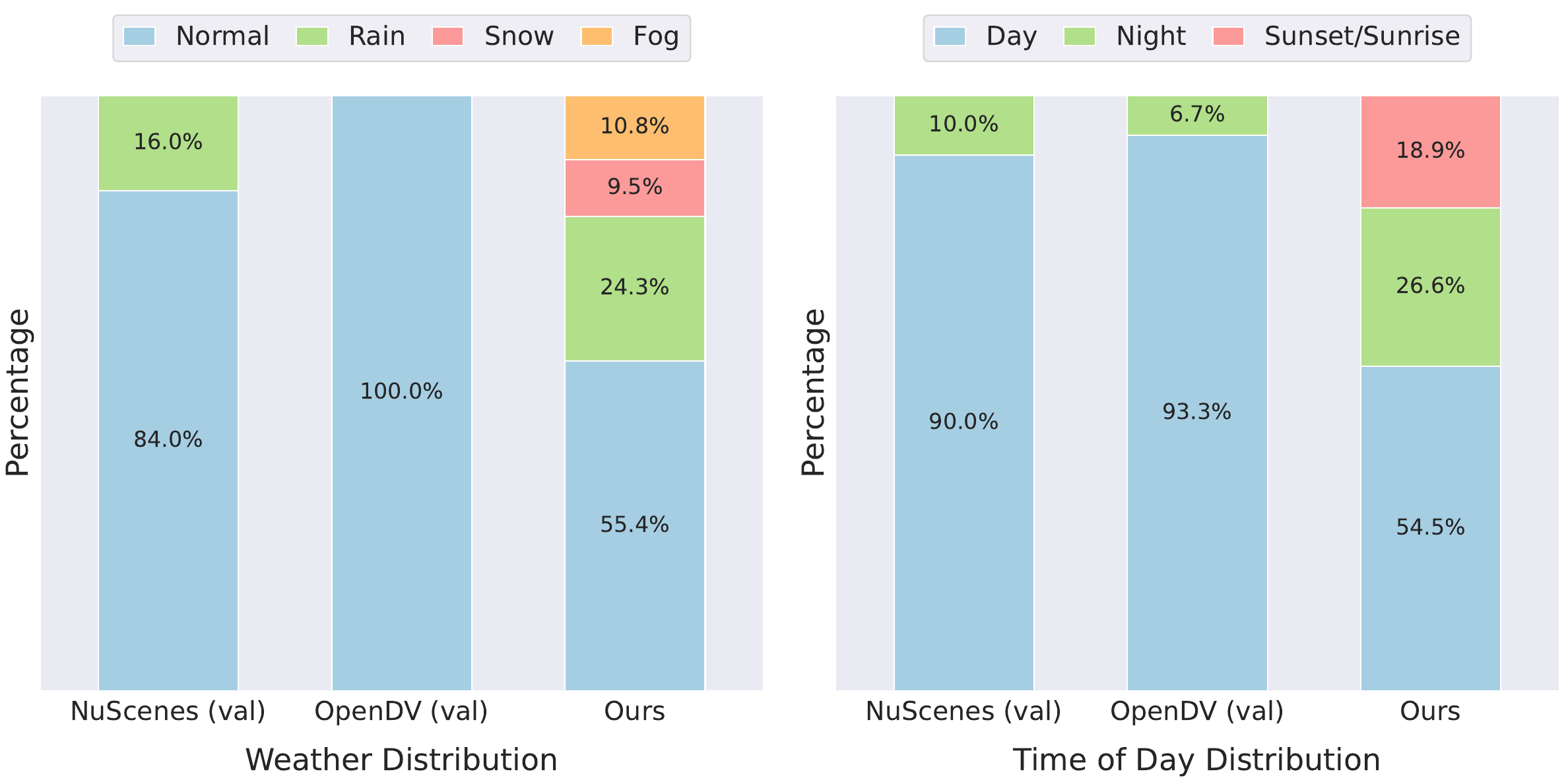}
    \caption{Weather and time of day distribution in our ego-condition track.}
  \end{subfigure}
  \hfill
  \begin{subtable}[t]{0.45\textwidth}
    \centering
    \resizebox{\linewidth}{!}{
      \begin{tabular}[t]{c|cc}
        \toprule
        Dataset Source & Region & Ratio \\
        \midrule
        Zod          & Europe & 26.5\% \\
        Drivingdojo  & China  & 25.6\% \\
        CoVLA        & Japan  & 27.4\% \\
        Nuplan       & U.S.   & 8.8\%  \\
        WOMD         & U.S.    & 11.6\% \\
        \bottomrule
      \end{tabular}
    }
    \caption{Data Ratio from existing open-sourced driving dataset in our ego-condition track.}
  \end{subtable}

  \caption{The statistics of our ego-condition track.}
  \label{fig:app_align_row}
\end{figure*}

\begin{figure}
    \centering
    \includegraphics[width=0.95\linewidth]
    {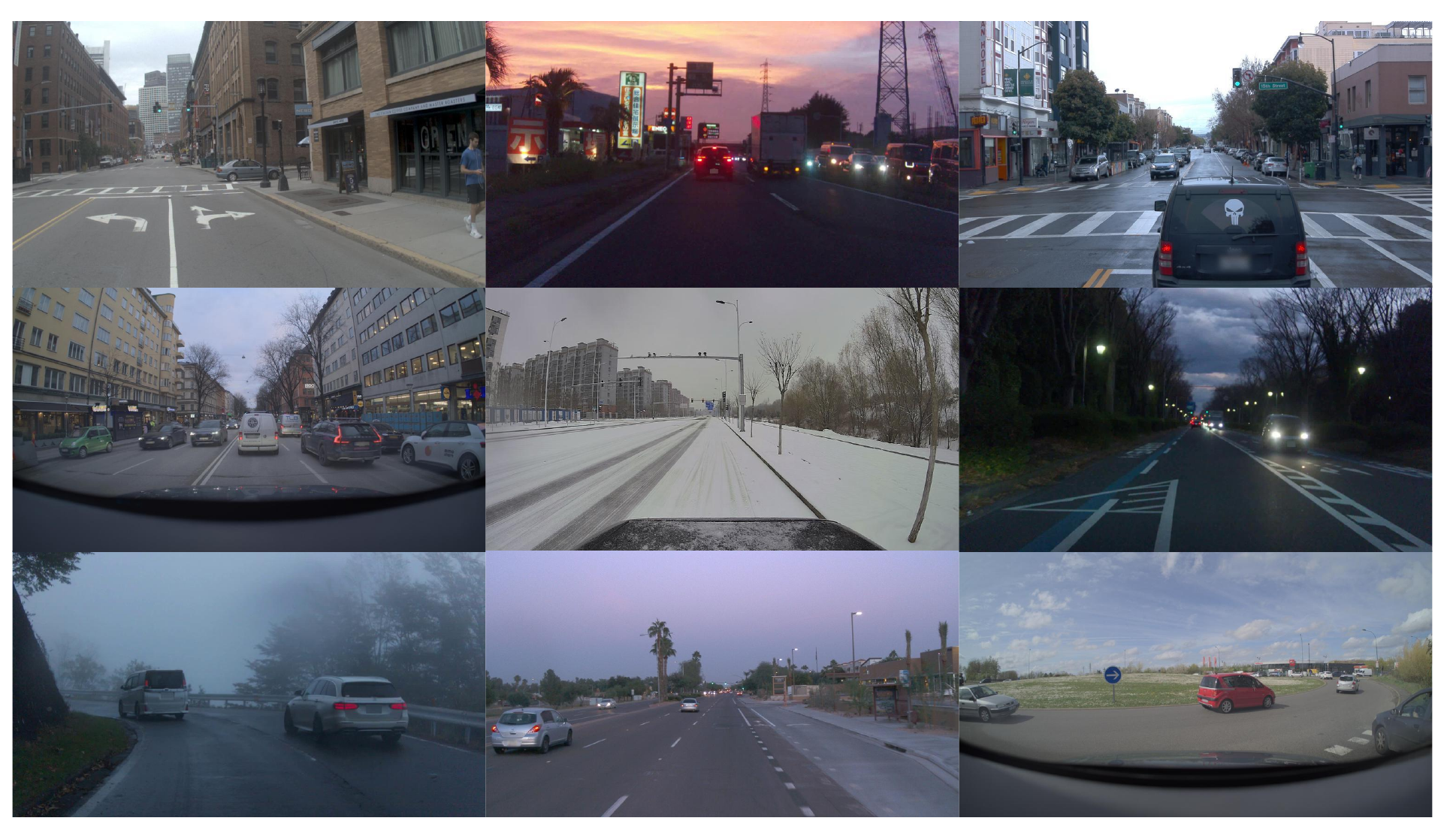}
    \caption{The gallery of our ego-condition track.}
    \label{fig:app_align_gallery}
\end{figure}

We present the distribution and gallery of our ego-conditioned track in Fig.~\ref{fig:app_align_row} and Fig.~\ref{fig:app_align_gallery}. We curated data from five open-sourced driving datasets to diversify the distribution of weather, time of day, and locations (with various driving styles). The videos and ego-trajectories provided in these datasets are used as the target distribution for calculating metrics such as FVD and FTD.

\subsection{\textcolor{black}{Details of Our SLAM Pipeline and compariision with Others}}
\label{app:slam_detail}

\textcolor{black}{\textbf{Dealing with Unsuccessful Trajectory Reconstruction.} Not every generated video will yield a successful SLAM reconstruction, especially if the video has tremendous artifacts or very low texture. Simply discarding those cases would bias the evaluation, because typically it’s the worst videos (the most unrealistic ones) that cause SLAM to fail. Dropping them would artificially inflate those poor-performing models’ scores. We tackled this issue explicitly to ensure no video is left unevaluated. Our approach was to build a custom SLAM+depth estimation pipeline that is robust to failures. We ensure a trajectory is obtained for every video by applying a failure-recovery strategy: if at any frame the SLAM algorithm cannot estimate the next camera pose (\textit{e.g.}, fails in feature matching, solving PnP, etc.), we take the last known pose and extrapolate it forward. Specifically, we propagate the last pose with a constant velocity model. To avoid giving an unrealistic advantage, we add small random perturbations to the pose orientation during this extrapolation. This injects a bit of uncertainty to mimic the fact that the current estimation is noisy, preventing the extrapolated path from appearing “too perfect” in our metrics. We chose not to simply freeze the camera (no movement) upon failure, because a completely static continuation could skew certain trajectory metrics. By using this continuous-and-jitter method, we obtain a complete trajectory from start to end for every video, no matter how poor its quality. This allows all videos to count toward the trajectory-based metrics, holding models accountable for cases where a naive SLAM would have given up.}

\textcolor{black}{\textbf{Comparison with Other SLAM Pipelines.}
We evaluated our reconstruction pipeline against those used in recent driving world-model systems.
Concretely, we compare the successful reconstruction rate and trajectory accuracy (ADE) on 20 nuPlan videos generated with Vista from our early experiments.
A run is counted as successful if the SLAM system returns a valid camera trajectory without numerical failure.
The results are summarized in Table~\ref{tab:slam_abl}. Compared to the GEM pipeline (DROID-SLAM~\citep{teed2022droidslamdeepvisualslam} + Depth-Anything v2~\citep{yang2024depthv2}) and the DrivinDojo pipeline (COLMAP~\citep{schoenberger2016mvs, schoenberger2016sfm} with scale aligned to ground truth), our basic version (\emph{Ours w/o failure handling}) achieves a similar successful reconstruction rate (17/20 vs.\ 17/20 and 16/20) and a comparable ADE (15.18 vs.\ 14.61 and 14.99).
When we enable our failure-handling strategy (\emph{Ours w/ failure handling}), the successful rate increases to 20/20, while the ADE remains in the same ballpark (16.84).
This trade-off is important for DrivingGen: the benchmark needs robust reconstruction on all videos rather than dropping harder cases and evaluating on a subset of “easy’’ videos.
Overall, our SLAM pipeline is more robust than existing pipelines by handling reconstruction failure explicitly.}

\begin{table*}[!t]
  \centering
  \begin{tabular}{lcc}
    \toprule
    Pipeline & Success rate $\uparrow$ & ADE $\downarrow$ \\
    \midrule
    GEM: DROID-SLAM + Depth-Anything v2   & 17 / 20 & 14.61 \\
    DrivinDojo: COLMAP + scale to GT      & 16 / 20 & 14.99 \\
    Ours w/o failure handling             & 17 / 20 & 15.18 \\
    Ours w/ failure handling              & 20 / 20 & 16.84 \\
    \bottomrule
  \end{tabular}
  \caption{\textcolor{black}{Comparison of different SLAM pipelines on 20 nuPlan videos generated with Vista.
    ``Success rate'' counts how many videos yield a valid reconstruction; ADE is the mean trajectory error over successfully reconstructed runs.}}
  \label{tab:slam_abl}
\end{table*}

\subsection{Fréchet Trajectory Distance (FTD)}
\label{app:ftd}

\textbf{Idea.} FTD applies the FID-style Gaussian Fréchet distance to \emph{trajectory} embeddings, replacing image/video features with a driving-domain encoder.

\textbf{Representation model and input.} We use MTR's \texttt{agent\_polyline\_encoder} $\phi(\cdot)$~\citep{shi2023motiontransformerglobalintention}. \emph{Crucially, MTR consumes a fixed temporal horizon $H$}.

\textbf{Window embeddings \& trajectory pooling.} We slice the trajectory into windows to fit into the MTR encoder. Each window is encoded as $\mathbf{f}=\phi(\text{window})\in\mathbb{R}^d$. A trajectory’s embedding is the mean over its window embeddings, which stabilizes statistics and removes dependence on the number of windows.

\textbf{Distributional distance.} For generated embeddings $X=\{\bar{\mathbf{f}}(\tau_i^{\text{gen}})\}_{i=1}^{n}$ and reference embeddings $Y=\{\bar{\mathbf{f}}(\tau_j^{\text{ref}})\}_{j=1}^{m}$ with empirical means/covariances $\hat{\boldsymbol{\mu}}_{X/Y}$, $\hat{\Sigma}_{X/Y}$, define
\[
\boxed{
\mathrm{FTD}(X,Y)=\|\hat{\boldsymbol{\mu}}_X-\hat{\boldsymbol{\mu}}_Y\|_2^2
+\operatorname{Tr}\!\left(\hat{\Sigma}_X+\hat{\Sigma}_Y
-2\big(\hat{\Sigma}_X^{1/2}\hat{\Sigma}_Y\hat{\Sigma}_X^{1/2}\big)^{1/2}\right)}
\]
We add $\varepsilon I$ ($\varepsilon{=}10^{-6}$) before the matrix square root and symmetrize products by $(A{+}A^\top)/2$ if needed. Optional Ledoit–Wolf shrinkage can be used when $n$ or $m<d$.

\textbf{Practical recipe (defaults).}
\begin{itemize}
  \item \textbf{Encoder:} MTR \texttt{agent\_polyline\_encoder}.
  \item \textbf{Horizon \& slicing:} $H{=}10$ steps; stride $s{=}H$ (non-overlapping); same slicing for generated and reference.
  \item \textbf{Normalization:} agent-centric translation/rotation per window; MTR schema constants $(\ell,w,h)=(4.5,2.0,1.8)$\,m; type=vehicle; validity=1.
  \item \textbf{Aggregation:} mean over a trajectory’s window embeddings; FTD on the two sets of trajectory-level embeddings.
\end{itemize}

\subsection{Objective Image Quality}
\label{app:obj_q}

\textbf{Motivation and background.}
Pulse–width modulation (PWM) in vehicle lighting and roadside luminaires induces temporal luminance modulation that, when sampled by rolling-shutter cameras, can alias into low-frequency flicker and degrade detection and tracking. The IEEE Automotive P2020 standard formalizes \emph{Modulation Mitigation Probability (MMP)} to quantify whether such modulation is sufficiently suppressed during operation~\citep{ieee2018ieee,ieee_p2020_draft_2022}. We implement MMP on the frame-mean luminance to provide a robust and efficient evaluation signal.

\textbf{Definition.}
Given frames $\{I_t\}_{t=1}^{T}$ at sampling rate $\textsf{fps}$, form the luminance sequence
$L_t=\operatorname{mean}(\mathrm{gray}(I_t))$ and its periodogram
$\widehat{P}(f)=\lvert\mathcal{F}\{L\}(f)\rvert^2$ (real FFT).
Let the dominant non-DC peak be
\[
f^\star \;=\; \arg\max_{\,f>0}\,\widehat{P}(f).
\]
If $f^\star<0.2\,\mathrm{Hz}$, set $\mathrm{MMP}=1$.

\textbf{Computation.}
With the band $B(f^\star)=\{f:\,|f-f^\star|<\Delta f\}$,
define the band-power ratio
\[
A \;=\; \frac{\sum_{f\in B(f^\star)} \widehat{P}(f)}{\sum_{f}\widehat{P}(f)+\varepsilon},
\qquad \varepsilon=10^{-8}.
\]
The metric is
\[
\boxed{\ \mathrm{MMP} \;=\; \mathbf{1}\!\left[A<\tau\right]\ }\in\{0,1\}.
\]

\textbf{Defaults.}
$\Delta f=\texttt{band\_hz}=0.5\,\mathrm{Hz}$,\;
$\tau=\texttt{thr}=0.05$,\;
$\textsf{fps}=10$.
The procedure uses a single FFT per clip with complexity $O(T\log T)$.

\subsection{Trajectory Quality}
\label{app:traj_q}

\textbf{Motivation.} Video-only scores can miss whether \emph{motions} are plausible and comfortable. We define a trajectory quality that aggregates three kinematic submetrics—comfort, motion, and curvature—via a weighted geometric mean (equal weights by default). Each submetric lies in $[0,1]$ with larger being better; we report per-trajectory scores and dataset means, skipping \texttt{NaN}s.

\textbf{Preliminaries.} A trajectory $\tau{=}\{(x_t,y_t)\}_{t=1}^T$. Velocities, accelerations, and jerks use centered finite differences. Heading comes from velocity, and yaw rate uses wrapped heading differences. Path length is the cumulative step distance. A trajectory is marked \emph{moving} if any speed exceeds $v_{\text{static}}{=}0.1$\,m/s.

\textbf{Comfort ($S_{\text{comf}}$).}
We score comfort from three per-meter peaks: longitudinal jerk, lateral acceleration, and yaw rate. Trajectories that are non-moving (speed $< v_{\text{static}}$) or too short ($\le 1$\,m) are set to \texttt{NaN}.
Each peak is then mapped to a \([0,1]\) component score with an inverse transform
\(S_q = 1/(1 + q/s_q)\) (higher is better), where \(s_q\) are scale factors
(default 1.0).
The final comfort score is the geometric mean of the three components.

\textbf{Motion ($S_{\text{speed}}$).} We penalize under-mobility using a trajectory’s mean speed. A monotone log mapping compresses high speeds and scales by $v_{\max}{=}k\,v_{\text{ref}}$ (defaults: $v_{\text{ref}}{=}6.0$\,m/s, $k{=}2.5$) to obtain $S_{\text{speed}}\in[0,1]$. Never-moving trajectories receive $0$.

\textbf{Curvature ($S_{\text{curv}}$).} Discrete curvature is formed from first/second derivatives of $(x_t,y_t)$. We then compute an RMS curvature $\kappa_{\mathrm{rms}}$, then map
\[
S_{\text{curv}}=\frac{1}{1+\kappa_{\mathrm{rms}}}\in(0,1].
\]
Non-moving trajectories return \texttt{NaN}.

\subsection{Agent Abnormal Disappearance}
\label{app:agent_missing}

\textbf{Motivation.} Agents should not vanish without a plausible cause (e.g., occlusion or leaving the view). We detect such cases directly from video with a minimal vision–language check.

\textbf{Method.} For each agent that disappears, we prepare \emph{three} frames: (1) the first frame where the agent is visible, (2) the last frame where it is visible (both with the agent box drawn in green), and (3) the first frame after it disappears (no box). We ask a VLM to classify the disappearance with the following prompt:

\begin{quote}\small
\texttt{Given three frames around the moment a green-boxed object disappears, classify the disappearance as \textbf{Natural} (e.g., occlusion or leaving the field of view) or \textbf{Unnatural} (abrupt or non-physical). Base your decision on visual and motion continuity and interactions with nearby objects. Output one word: Natural or Unnatural.}
\end{quote}

\textbf{Scoring.} A tracklet is \emph{abnormal} if the VLM outputs \texttt{Unnatural}; otherwise it is \emph{not abnormal}. A video is \emph{clean} only if all evaluated tracklets are \emph{not abnormal}. The final score is the percentage of clean videos (higher is better).

\begin{table*}[!t]
  \centering
  \begin{tabular}{ccc}
    \toprule
    Component & Example & Approx.\ Time \\
    \midrule
    \multirow{2}{*}{Video Generation} & Wan2.2-14B & Days \\
    & Vista & About One Day \\
    \midrule
    Trajectory Reconstruction &
    SLAM + Depth model &
    Hours \\
    \midrule
    \multirow{2}{*}{Distribution Metrics} &
    FVD & Hours \\
    & FTD & Minutes \\
    \midrule
    \multirow{3}{*}{Quality Metrics} &
    Subjective Image Quality & Hours \\
    & Objective Image Quality & Minutes \\
    & Trajectory Quality & Minutes \\
    \midrule
    \multirow{4}{*}{Consistency Metrics} &
    Video Consistency & Hours \\
    & Agent Consistency & More Hours \\
    & Agent Disappearance Consistency & Hours \\
    & Trajectory Consistency & Minutes \\
    \midrule
    \multirow{2}{*}{Trajectory Alignment Metrics} &
    ADE & Minutes \\
    & DTW & Minutes \\
    \midrule
    All Metrics (Total) &
    All Above Metric Groups &
    1--2 Days on a Single GPU \\
    \bottomrule
  \end{tabular}
  \caption{\textcolor{black}{Approximate runtime of different components in DrivingGen on 400 videos with 100 frames each, evaluated on a single modern GPU. Times are coarse estimates and may vary with hardware.}}
  \label{tab:metric_time}
\end{table*}

\subsection{Trajectory Consistency}
\label{app:traj_c}

\textbf{Definition.} From positions sampled at step $\Delta t$, form the speed series $v_t$ and the acceleration series $a_t$ by finite differences. Measure each signal’s dispersion relative to its typical level using a simple ratio, then squash with an exponential:
\[
R_v=\frac{\mathrm{std}(v)}{\mathrm{mean}(v)},\quad
R_a=\frac{\mathrm{std}(a)}{\mathrm{mean}(|a|)},\qquad
S_v=\exp(-R_v),\ \ S_a=\exp(-R_a).
\]
The trajectory consistency score is the average
\[
S_{\text{cons}}=\tfrac{1}{2}\,(S_v+S_a)\in(0,1],
\]
where higher indicates smoother, more realistic kinematics.

\subsection{\textcolor{black}{Time and Resource for DrivingGen}}
\label{app:time}

\textcolor{black}{In our experiments, the bottleneck is primarily the video generation itself: many of the state-of-the-art generative models we benchmark are slow and memory hungry (e.g., Wan2.2-14B takes about 20-30 minutes to generate one 100-frame video on a single GPU with at least 40 GB memory). In contrast, the evaluation suite is comparatively manageable. The approximate wall-clock time for each metric group on 400 videos is summarized in Table~\ref{tab:metric_time}. On a single modern GPU, running all metrics for 400 videos with 100 frames takes roughly 1–2 days.}

\textcolor{black}{Within this budget, the main cost on the evaluation side comes from image quality and video consistency metrics, which require running heavy visual backbones over every frame. The most time-consuming metrics would be agent consistency and disappearance consistency, which run models for each agent in the first frame of the video. Trajectory measures (FTD, quality, consistency and alignment) are much cheaper (minutes), since they operate on compact embeddings or low-dimensional trajectories. These numbers are indicative and may vary with hardware and implementation, but they show that: (i) video generation dominates the overall runtime, and (ii) among the metrics, the image, video and agent quality and consistency components are the main contributors, while the rest of the metrics are comparatively fast.}

\subsection{Human Alignment of DrivingGen}
\label{app:human_align}

We employ a similar method in VBench to determine whether each category aligns with human preferences. Given the human labels, we calculate the win
ratio of each model. During pairwise comparisons, if a
model’s video is selected as better, then the model scores
1 and the other model scores 0. If there is a tie, then both
models score 0.5. For each model, the win ratio is calculated as the total score divided by the total number of pairwise comparisons in which it participated. 

\begin{figure}
    \centering
    \includegraphics[width=0.95\linewidth]
    {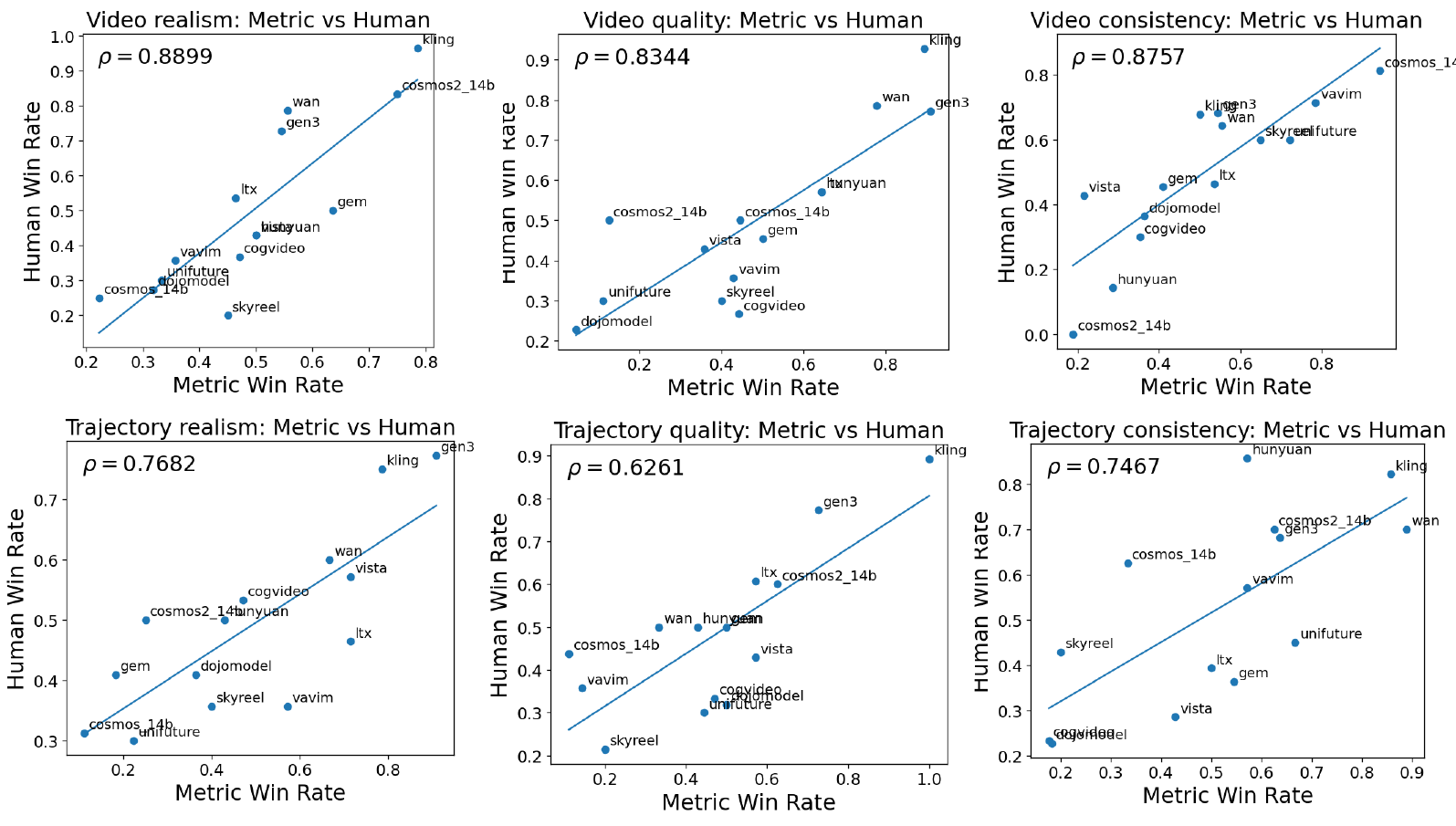}
    \caption{Human Validation of Our benchmark. Our metrics closely match human preferences. Trajectory-related metrics are less accurate in comparison to humans, likely due to noisy monocular SLAM and metric-depth recovery from generated videos with artifacts.}
    \label{fig:human}
\end{figure}

For fast and reasonable evaluation, we select three categories: distribution, quality and consistency. We evaluate with both videos and trajectories and use the primary metric in each category. Metrics are FVD and FTD, Subjective image quality and trajectory quality, video consistency and trajectory consistency. The results are shown in Fig.~\ref {fig:human}.

\end{document}